# Brain Connectivity Impairments and Categorization Disabilities in Autism: A Theoretical Approach via Artificial Neural Networks

**Daniele Q. M. Madureira**
Laboratório Nacional de Computação Científica
Coordenação de Matemática Aplicada e Computacional
Av. Getúlio Vargas, 333, Petrópolis, RJ, Brasil

**Vera Lúcia P. S. Caminha**
Universidade Federal Fluminense
Instituto de Ciências Exatas
Laboratório ADACA
R. Des. Ellis Hermydio Figueira, 783, Volta Redonda, RJ, Brasil

**Rogerio Salvini†**
Universidade Federal de Goiás
Instituto de Informática
Alameda Palmeiras, Quadra D, Campus Samambaia, Goiânia, GO, Brasil

**Abstract**
**Background:** A developmental disorder that severely damages communicative and social functions, the Autism Spectrum Disorder (ASD) also presents aspects related to mental rigidity, repetitive behavior, and difficulty in abstract reasoning. According to neuroscientific studies, imbalances between excitatory and inhibitory brain states, in addition to cortical connectivity disruptions, are at the source of the autistic behavior.
**Objective:** As the exact relationships between such brain impairments and the cognitive difficulties in the ASD still remain poorly understood, our main goal consists in unveiling the way by which these local excitatory imbalances and/or long brain connections disruptions are linked to the above mentioned cognitive features.
**Methods:** We developed a theoretical model based on Self-Organizing Maps (SOM), where a three-level artificial neural network qualitatively incorporates these kinds of alterations observed in brains of patients with ASD. Although previous works based on SOM networks addressed autistic features as islands of abilities and attention focusing on details, in this work, we provide a novel and more realistic approach that presents a direct relationship between cortical features maps elements and both synaptic and axonal characteristics of the utmost importance for the comprehension of the neural basis of autistic cognition.
**Results:** Computational simulations of our model indicate that high excitatory states or long distance under-connectivity are at the origins of cognitive alterations, as difficulty in categorization and mental rigidity – both possibly linked to the autistic repetitive behavior. More specifically, the enlargement of excitatory synaptic reach areas in a cortical map development conducts to low categorization (over-selectivity) and poor concepts formation. And, both the over-strengthening of local excitatory synapses and the long distance under-connectivity, although through distinct mechanisms, contribute to impaired categorization (under-selectivity) and mental rigidity.
**Conclusions:** Our results indicate how, together, both local and global brain connectivity alterations give rise to spoiled cortical structures in distinct ways and in distinct cortical areas. These alterations would disrupt the codification of sensory stimuli, the representation of concepts and, thus, the process of categorization - by this way imposing serious limits to the mental flexibility and to the capacity of generalization in the autistic reasoning.

**† Corresponding author:** Rogerio Salvini, Universidade Federal de Goiás, Alameda Palmeiras, Quadra D, Campus Samambaia, Goiânia, GO, Brasil - CEP 74690-900. E-mail: rogeriosalvini@inf.ufg.br. Phone number: +55 62 3521-1181 . Fax number: +55 62 3521-1182.

1. **Introduction**

In the ASD, complex brain alterations markedly affect both communication and social abilities. Moreover, everyday activities are hard to achieve due to the interference of repetitive behavior patterns, mental rigidity and categorization difficulties (Rogers and Pennington, 1991), (Minshew *et al*., 2002), (Hill and Frith, 2003), (Mundy, 2003), (Happe and Frith, 2006), (Goldstein *et al*., 2008), (O´Hearn et al., 2008), (Geschwind, 2009), (Lewis and Kim, 2009), (Zikopoulos and Barbas, 2010), (Pelphrey *et al*., 2011), (Kana *et al*., 2011), (Boucher, 2012) and (American Psychiatric Association, 2013).

The joint effort of several medical areas tries to understand the complexity of the autistic symptoms. Indeed, multidisciplinary teams of neurologists, psychiatrists, psychologists, speech therapists and pedagogues usually work together for the welfare of patients with ASD. Also, inspired by the ASD supporting tools currently used by such health professionals – i.e., the Picture Exchange Communication System (PEC) (Frost and Bondy, 1994), (Flippin *et al*., 2010), or the Applied Behavior Analysis (ABA) (Johnston, 2015), (Lotan *et al*., 2015) – the computer science community started creating digital games with the purpose of improving cognitive abilities, or enhance the interactivity of ASD patients (Passerino et al., 2007), (Caminha *et al*., 2013). Even so, the relationship between neurobiological substrates and the diversity of cognitive symptoms observed in ASD remains poorly understood. This is why there are currently numerous studies in neuroscience, still struggling to understand the specificities of this developmental disorder.

The variety of ASD behavior alterations and the distinct levels of neural processing give rise to quite different neuroscientific approaches for scrutinizing autistic features. For instance, research on mirror neurons attempt to address the lack of empathy or theory of mind difficulties in the ASD (Iacoboni, 2009), (Libero *et al*., 2014), (Mostofsky and Ewen, 2011). Alternatively, studies involving cerebral rhythms deal with synchrony between brain areas, analyzing how altered neural rhythms affect the propagation of information throughout the brain (Bosl *et al*., 2011), (Lazarev *et al*., 2010, 2015). Another research strategy consists in accessing connectivity patterns around the brain. In particular, many studies have been revealing axonal integrity alterations, which harm the neural signaling of long distance projections (Courchesne and Pierce, 2005), (Geschwind and Levitt, 2007), (Zikopoulos and Barbas, 2010), (Kana *et al*., 2011), (Mostofsky and Ewen, 2011), (Schipul *et al*., 2011). For instance, Courchesne and Pierce (2005) proposed that impairments in both frontal-posterior and posterior-frontal connections underlie the lack of synchronization in autistic brains; moreover, Geschwind and Levitt (2007) pointed to disconnections in frontotemporal couplings and in callosal tracts; and Kana *et al*. (2011) referred to a frontoparietal under-connectivity.

Although distinct, such approaches admit that an essential aspect underlies the ASD symptoms: the functional integration disruption between cortical areas (Schipul *et al*., 2011), (Kana *et al*., 2011).

In this work, we explore qualitatively the *connectivity patterns paradigm*, through Self-Organizing Maps (SOM), a type of artificial neural network (Kohonen, 1982), (Haykin, 1999), to model and simulate the development of cortical maps. We speculate how, in autistic brains, physiological imbalances in local synapses (Rubenstein and Merzenich, 2003), (Gogolla *et al*., 2009) and/or the malformation of long projections (Courchesne and Pierce, 2005), (Geschwind and Levitt, 2007), (Kana *et al*., 2011), possibly hamper, or even prevent, the process of categorization that is essential for a satisfactory functioning of the cognitive apparatus (Minshew *et al*., 2002). More specifically, an impaired neural substrate for categorization would be directly related to the mental rigidity and the repetitive behavior frequently manifested in ASD patients (Zandt *et al*., 2007), (Lewis and Kim, 2009).

Throughout the neurodevelopment of cortical feature maps - brain structures that codify representations of sensory stimuli - well defined neuron groups become responsive to specific classes of signals. On account of its topographic structure, a cortical feature map presents properties directly related to selectivity and categorization. The nearer a neuron is in relation to another, the more similar the patterns these neurons codify. As a consequence, in a map, a particular neuron cluster codifies a particular class of stimuli. In addition, a stimulus that is more frequently presented to the map originates a larger neuron cluster to codify it.

According to our approach, such maps are essential parts of the neurobiological substrate of the categorization process, which is fundamental for the capacity of making generalizations.

SOM networks incorporate cortical feature maps properties. In particular, those ones that appear due to mechanisms of competition and cooperation between neurons, which occur during a map development. Basically, competition relies on the inhibition that a particular neuron exercises on distant neurons activity, while cooperation depends on the excitatory connections between neighbor neurons Carvalho *et al*. (2003), (Mendes *et al*. (2004) and Gustafsson (1997). So, by simulating self-organizing properties underlying the development of cortical maps, SOM networks allow investigations on how sensory information is topographically codified in cortical areas. Moreover, it conveys a biological inspired computational technique to infer how the integration of distinct levels maps possibly leads to the emergence of abstract information, since it is plausible that concepts are organized in semantic cortical maps (Ritter and Kohonen, 1989).

The application of SOM networks for modeling brain mechanisms has already provided interesting insights into neuropsychiatry. With relation to the ASD, the computational model and simulations proposed by Carvalho *et al*. (2001) linked alterations in levels of neural growth factors to disruptions in cortical maps and the emergence of islands of ability – a typical manifestation in ASD. Also, the autistic weak central coherence – or cognitive focusing on details – was addressed by Gustafsson (1997) and by Noriega (2007). And, recently, Noriega (2015) proposed a model for investigating sensorial processing alterations in ASD. See (Cohen, 1994), (O'Loughlin and Thagard, 2000), (Klin *et al*., 2003), (Happe and Frith, 2006), for other modeling approaches in ASD.

Also, a continuum of mental states ranging from mental rigidity up to the schizophrenic disorganized mind, passing through both the normal and creative thinking, was also investigated via SOM (Carvalho *et al*., 2003), (Mendes *et al*., 2004). In this case, the neurocomputational model illustrated how dopamine interferes in the cortical signal-noise rate, thus inducing different mental states – normal or pathological ones.

From these SOM approaches, here we address phases in the development of cortical maps (Buonomano and Merzenich, 1998), (Mendes *et al*., 2004), to form a qualitatively–plausible view of how the excitatory synaptic imbalance in autistic brains suggested by Rubenstein and Merzenich (2003) and/or the weakening of long distance projections (Courchesne and Pierce, 2005), (Geschwind and Levitt, 2007), (Zikopoulos and Barbas, 2010), (Kana et al., 2011) – also in ASD patients – interfere in the organization of cortical maps, i.e., in the neural codification of conceptual classes.

Here, through a multilayer neural circuit composed of three hierarchical levels, we investigate if and how disruptions in some level of the network cause further damages on higher levels maps, since a significant body of studies reveal that patients with ASD present significant difficulties in activities involving theory of mind, working memory, response inhibition and planning - i.e., higher cognitive, executive functions that depend on high order cortical processing (Kana *et al*., 2011), (O῀Hearn *et al*., 2008).

Summarizing, we propose that the topographically organized neural encoding of conceptual classes depends on biological mechanisms as neural competition and cooperation, and relies on a balance between synaptic excitatory and inhibitory influences. Keeping in mind the lack of balance between local excitatory and inhibitory connections in autistic brains, as indicated by Rubenstein and Merzenich (2003), and the lack of inhibitory control suggested by Courchesne and Pierce (2005), it is plausible to question if such alterations do disrupt the neural substrates for categorization in ASD (the topographically organized cortical feature maps). We suspect that a pathological process, in the ASD, could develop neural structures that are not biological substrates for the cognitive capacity of categorizing, as we believe it is the case with the normal cortical feature maps. In addition, by qualitatively modeling a neural circuit composed of distinct maps, we expect to proceed towards a better theoretical view of the lack of integration between distinct brain areas, in particular, the alterations in long connections observed in autistic brains (Courchesne and Pierce, 2005), (Geschwind and Levitt, 2007), (Zikopoulos and Barbas, 2010), (Kana et al., 2011).

Therefore, by indirectly modeling neurobiological characteristics previously described as related to the ASD in numerous neuroscience studies - in particular, excitatory-inhibitory states imbalance and axonal disruptions -, our goal is to delineate a link between inter-related autistic features - such as difficulty in dealing with categories, mental rigidity, and repetitive behavior -, and the way neurons self organize and transmit signal in autistic brains.

Finally, it is worth remarking that, here, the SOM paradigm is implemented to simulate a brain mechanism associated to the ability of forming categories. As an application in a theoretical neuroscience study, the SOM algorithm is employed to depict the neural process that we are interested in investigating. Additionally, throughout this work we refer to our model as qualitative, because despite dealing with numerical quantifications, the amounts are not related to measures directly taken experimentally.

The rest of this paper is organized as follows: in Section 2, we describe the modeling details: the network architecture and dynamics, the input patterns construction, and the simulations designs. In Section 3, we present the results of our simulations. In particular, we explore the *local excitatory imbalance*, through synaptic over-strengthening or excitatory area enlargement and the *disruption of long distance connections*. We also delineate an association between the simulated maps states and some autistic cognitive features  In Section 4, we discuss our model and simulations. Finally, in Section 5, we summarize our conclusions.

## 2. Proposed Model

We propose here a minimal multilayer neural circuit, composed of three hierarchical levels, in order to qualitatively simulate the propagation of neural signals from two primary cortical areas up to a higher one, passing through an associative region.

Essentially, our main idea resides in investigating neurobiological substrates for categorization skills, especially in autistic brains, under a theoretical approach. Accordingly, we observe the development of each map in the network, due to the presence of a particular input set, both in the normal condition and under autistic modeled neural alterations.

### 2.1. Network Architecture and Dynamics

In the ASD, the cognitive processing of concrete features seem to be less impaired than the abstract reasoning, and experimental studies indicate disruptions between associative and higher order cortical areas. So, we propose a minimal architecture topology, with cortical levels enough for us to start investigating if or how the integration and posterior propagation of lower level signals - typically associated to concrete objects - become disrupted whenever the above mentioned autistic neural alterations are present, thus preventing higher cortical regions - usually active in processes involving abstraction - from achieving an adequate functioning.

Our multilayer network is thus composed of three levels: a primary one, with two maps (*Prim1* and *Prim2*), that represents a cortical processing relative to sensory, concrete stimuli of distinct types. In the second layer, there is a single map (*Assoc*) that processes the integration of signals coming from the first level. This map represents an associative cortical area. And, in the third level, another single map (*Front*) represents a higher cortical region. Therefore, through this three-level approach we are able to address a bottom-up pathway departing from neural signals directly related to concrete information, up to a more abstract representation in a higher cortical area. Figure 1 depicts the modeled neural network.

Preprocessed sensory stimuli reach the primary maps, *Prim1* and *Prim2*. At this level, each map processes signals that correspond directly to a particular kind of sensory information, i.e. visual, auditory or somatosensory ones. In our model, thus, *Prim1* and *Prim2* symbolize different types of cortical maps, which codify distinct sensory stimuli.

Projections that leave *Prim1* and *Prim2* reach the *Assoc* map. As we note in Figure 1, inputs to the second level map are composed by the integration of *Prim1* and *Prim2* outputs. This signal is a combination of distinct maps outputs. Therefore, at this level, neural information lacks its direct correspondence with sensory stimuli, appearing as abstract signals.

The output of the *Assoc* map goes straight to the *Front* map – at the network last level. There, the neural signal carries an even higher level of abstraction.

## 2.2. Normal Maps Development

In our model, each map undertakes a learning process until it reaches a final state - a developed map, where neuron clusters codify the classes of stimuli that were presented to the network. According to the SOM learning algorithm, an input pattern is randomly chosen from some input set, and activates each neuron in the map. The activation degree depends both on the input value and the weight of the connection between the neuron and the input pattern. The neuron with the highest activation value is called *winner*. This is the competitive section of the learning process. Next, there occurs the cooperative step, where it takes place the updating of weights. The *winner* and its neighboring neurons have their connections weights strengthened, so that they become more responsive to the stimuli class just presented to the network.

This process happens repeatedly until the map reaches a stable state (Kohonen, 1982), (Haykin, 1999), (Carvalho et al., 2001) and (Carvalho et al., 2003). By this way, neurons pertaining to a given neighborhood develop quite similar weight connections, which contribute to their simultaneous activation, when patterns from a specific class reach the cortical map. In other words, for each input class, there will be a well defined cluster of neurons – the encoding neurons – which responds to the arrival of any stimulus from its corresponding class.

As a consequence, a well-organized, normal map presents the following characteristics: its number of neuron groups matches the number of classes of its input set and, there is no superposition in neuron coding (in other words, a neuron does not respond to stimuli from distinct classes). And, for both normal or pathological conditions, the smaller a neuron cluster is, the smaller the representation of the corresponding class - or, the poorer the concept representation; and, the larger a neuron cluster is, the larger the cortical region assigned to process its respective stimuli class (Carvalho et al. (2001) and Carvalho et al. (2003)).

### 2.3. Input Patterns

With respect to the network first level, there are two distinct input sets – one for *Prim1*, and another for *Prim2*. Each input set contains four patterns, composed by randomly generated points belonging to each of the four quadrants ([0,10] x [0,10], [0,10] x [10,20], [10,20] x [0,10] and [10,20] x [10, 20]) in the domain [0, 20] x [0, 20] ⊆ $R^2$ . The purpose of such construction is, under normal conditions, to achieve basal maps - each one with four defined neuron clusters thus reflecting the input sets organization. Or, simulate a neural self-organizing mechanism that is able to form a neurobiological structure topologically related to the input patterns: a cortical feature map.

Once *Prim1* and *Prim2* are developed, input stimuli to the *Assoc* map are shaped. Initially, a bi-dimensional coordinate is randomly chosen from the encoding neurons in *Prim1*, and another from *Prim2*. Next, they are concatenated, thus forming a four-dimensional pattern. The input set to *Assoc* is so generated, and is composed by four patterns – the same size of the input sets to *Prim1* and *Prim2*.

Propagation of neural information to the *Assoc* map depends on which neurons become activated in *Prim1* and *Prim2*. Since an organism processes simultaneous, distinct stimuli, each input pattern to *Assoc* represents both the simultaneity and the randomness that characterizes the interaction between an organism and its environment.

Again, once *Assoc* reaches a stable state, the input set to the *Front* map is generated. In this model, the *Front* map receives projections from the *Assoc* map only. Therefore, its input set is composed by randomly chosen coordinates that identify the positions of encoding neurons in the *Assoc* map. As in the previous level, the input set to the *Front* map contains four patterns.

### 2.4. Local Excitatory Imbalance

As previously exposed, studies indicate, for autistic brains (Rubenstein and Merzenich, 2003), some imbalance between excitatory and inhibitory connections, leading to an increase in the excitatory state. Here, we describe how a local excitatory alteration is modeled via our SOM network.

In SOM, self-organization depends on short distance excitatory synapses and long distance inhibitory ones (Kohonen, 1982). So, we investigate the effects of enlarging excitatory parameters during the network learning process: we raise both the synaptic strength and the excitatory area.

*Excitatory Synapses Over-Strengthening.* According to the SOM learning algorithm, when a stimulus $x_k$ reaches the network, the synaptic updating, $\Delta w_i$, for the i-th neuron in the map, obeys the following steps:

$$w_i = w_i + \Delta w_i,$$
$$\Delta w_i = \rho.\varphi(r_i,r^*).(x_k - w_i), \quad (1)$$

where $\rho$ is the learning rate, $\varphi$ is the neighboring function, and $r_i$ and $r^*$ are the positions of the neuron i and the winner neuron in the map.

Here, we model an increase in the lateral (short-range) excitatory effect (synaptic weight) by augmenting the amount by which the neurons that are neighbors to the winner neuron have their weight changed, as:

$$w_i = w_i + \Omega.\Delta w_i, \quad (2)$$

where $\Omega$ is the exceeding amount due to the excitatory unbalance.

*Excitatory Area Enlargement.* In this case, we improve the excitatory influence area throughout the map development. In terms of the SOM learning algorithm, the Gaussian function $\varphi(r_i,r^*)$ defines the winner neuron

neighborhood:

$$\varphi(r_i, r^*) = \exp\left(-\|r_i - r^*\|^2 / 2\sigma^2\right) \quad (3)$$

So, the i-th neuron has its weight, $w_i$, updated depending on its position, $r_i$, in the map, relative to the winner neuron position $r^*$. Or, if the i-th neuron pertains to the winner neighborhood, $\varphi(r_i, r^*)$, it gets the right to strengthen the connections that provide information from the stimulus, $x_k$. The parameter $\sigma$ regulates the neighborhood extension. Therefore, we modeled excitatory area enlargements by imposing variations in its value.

### 2.5. Disruption of Long Distance Connections

A number of neuroscience studies indicate disruptions in long distance projections in autistic brains. In particular, they reveal under-connectivity between frontal and parietal areas (Kana *et al.*, 2011), and disconnections between frontal and temporal regions (Geschwind and Levitt, 2007).

Then, in this section, we consider axonal impairments that corrupt connections between distinct cortical areas. More specifically, here, we address the axonal under-connectivity. Once long neural projections are nonfunctional, integration of distinct kinds of neural information is affected, and so is the cognitive process.

The connections between maps, described in Figure 1, represent axonal projections that allow the traveling of neural signals from one cortical region to another. Integration and synchronization of different neuron groups are of essential importance during brain activities. And, such processes depend on precise arrival times of neural signals at the synapses. Thus, disruption of long connections may have devastating effects, as in the ASD case.

*Impairment of Axonal Projections.* Here, we address the situation where connectivity damages lead to the weakening of neural signals. Or, in artificial neural networks terms, the connection´s weight enfeebles due to axonal disruption, thus decreasing the activation of the pos-synaptic neuron.

Throughout the learning phase, network weights acquire final, suited values. In a SOM network, such values are responsible for the network capacity of organizing clusters. In Equation (1), we note that weights updating depends on various factors, as learning rate, winner neighboring, and distance between a given neuron and the winner one.

We model a pathological neurodevelopment where connections do not reach an ideal weight. In this case, we represent it through the weakening of the learning rate $\rho$. By this way, we indirectly represent a delay in the axonal formation. It thus prevents the axonal projections to acquire - at the right time - an elaborated structure, fundamental for the neural signal maintenance and transmission.

### 3. Simulations Results

The modeled network was implemented in the ANSI C® programming language on the basis of the characteristics presented above. We performed a series of experiments, departing from what was defined as a well-organized, normal map, up to the modeled pathological situations relative to connectivity alterations.

Firstly, we simulated the normal development of cortical maps in a three-level neural network. In the sequence, we imposed some local pathological conditions in single maps. Only then, in the following experiments, we started observing the effect of local impairments in distant areas. Finally, we accessed both the local alterations and the long projections disruptions in the three-level network. Table 1 presents the gauged parameters used.

| Maps dimension | 20 X 20 |
|---|---|
| Learning rate (ρ) | 0.5 |
| Neighborhood constant (σ) | 2.0 |
| Encoding threshold (Θ) | 0.999 |
| Learning steps | 1400 |

**Table 1.** Default parameters.

### 3.1. Normal Condition

In our first simulation, as shown in Figures 2a-d, each primary map reaches an organized final state, with well characterized clusters of encoding neurons that reflect the input sets contents. The default parameters are described in Table 1. Since no pathological alterations are present, the maps capacity of representing sensory categories remains intact. In the network second level, the *Assoc* map also manifests a well shaped organization in clusters. Here, neurons codify an integration of signals. So, although abstract, the neural information remains classified. Finally, as expected, the *Front* map encoding neurons are organized in clusters too, thus reflecting the input set structure.

This simulation provided us a referential network. Here, all maps qualitatively represent neurobiological substrates for the cognitive ability of forming categories.

### 3.2. Local Excitatory Imbalance

In the sequence, we explore the problem of local excitatory arousing during a limited stage in the neurodevelopment. In this Section, we will look at single maps, isolated from the network.

***Excitatory Synapses Over-Strengthening.*** In the first situation, weights suffer an over-updating that strengthens them three times as much as in the normal condition, with $\Omega = 3$ in Equation (2). As a result, exemplified in Figures 3a-d, the map does not achieve a stable state. Just the opposite, during its development, transient clusters continuously turn into another ephemeral groups. In cognitive terms, this situation could be associated to no concepts consolidation.

When the weights over-updating is even more exacerbated, i.e. five times as much as in the normal condition - with $\Omega = 5$ in Equation (2) -, the organization in clusters does not occur anymore. Moreover, most of the neurons do not become encoders. We note in Figure 4 that only a few neurons have the chance to be activated: no neural structure at all supports the construction of categories. Identical results were attained when we applied an even higher increase factor, $\xi(r_i, r^*)$, for weights updating, as:

$$\Delta w_i = \rho \cdot \xi(r_i, r^*) \cdot (x_k - w_i),$$

where

$$\xi(r_i, r^*) = \varphi(r_i, r^*)/(C - \varphi(r_i, r^*)), \quad (4)$$

with the constant C set as 1.1.

***Excitatory Area Enlargement.*** In this second approach, we make a set of successive simulations where the excitatory synapses updating area is gradually increased, that is, in each simulation there is an increase in the parameter σ, in Equation (3), which controls the neighborhood area of winner neurons during the learning process.

When σ = 0.5, a low value, the map does not reach an organized, clustered final state, and remains disorganized. Differently, for higher values of σ, such as 1.0 and 1.5, groups of encoding neurons start to appear.

These maps, however, are not completely organized, and a number of neurons are out of their clusters.

The next simulation provides a stable, well organized final map. Here, σ is set as 2.0, and refers to a normal condition - cf. with Figure 2a. In the sequence, we impose a further increase in σ, which assumes the value of 3.0. This map reaches a well organized configuration, but its neuron groups are more distant from each other, if compared with the normal condition shown in Figure 2a. Interestingly, according to Carvalho et al. (2003), in the above mentioned work on mental processes via SOM, the enlargement of distances between neuron clusters would indicate a decrease in the capacity of making associations, typical of a less flexible reasoning. Although there are slightly less encoding neurons (from 251, for σ = 2.0, to 236 - a decrease of 5.98%), this map keeps an organization considered normal, as described in Section 2.2.

In the sequence, for σ values of 4.0 and 5.0, the excitatory area becomes even more enlarged, and the clusters come to be even more distant from each other, and the number of encoding neurons drops drastically to 149 and 73, respectively - see in Figure 5a the map developed with σ = 5.0. In cognitive terms, such a neuron groups shrinkage signalizes a neurobiological structure incompatible with the processing of generalization.

At last, when σ assumed the values of 6.0, 7.0 and 8.0, a novel feature emerges. Besides the enormous distance between the clusters, and the low number of encoding neurons in each map (29, 28, 17, respectively), there appears a superposing of neural codification. Although the input set contains four patterns, distributed in distinct quadrants, the final map presents only three neuron groups. Distinct stimuli are encoded by the same cluster. It thus, plausibly, suggests a lack of selectivity in the cognitive processing, besides a great difficulty in making associations. Figures 5b,c depict the final states of maps where σ is set as 7.0 and 8.0, respectively.

**3.3.    Local Impairments Influence on Higher Level Map**

The results just presented in Section 3.2 suggest that an exceeding strengthening of excitatory synapses prevents a single map from achieving an adequate development, i.e. a topographical organization in clusters. Thus, our next step was to investigate consequences of such underdevelopment on the following higher level map.

Here, we use a two-level network, composed by *Prim1* and *Prim2* at the first level, and *Assoc* in the second level. The connections weights of *Prim1* and *Prim2* were updated through the exceeding amount, ξ($r_i$, r*), as described in Equation (4), with C set as 1.1. The final state can be seen on Figures 6a, b.

Since neural signaling to a higher level map depends on encoding neurons, we expected some aberrant situation due to the absence of development of *Prim1* and *Prim2*. In fact, the tiny amount of encoding neurons in *Prim1* and *Prim2* enhanced the appearance of quite similar patterns in the input set to *Assoc*. Therefore, this higher level map developed under the abnormal repetition of fixed signals. As we observe in Figure 6c, there appear only two neuron groups. One of them is responsible for responding to a particular signal (marked with asterisks "*"). The other, however, codifies three kinds of signals simultaneously, because they are very similar.

Thus, in comparison to the normal condition, the excitatory imbalance in the first level maps led to: diminishing in the number of neuron clusters, enlargement of the encoding areas, and superposition of codification, in the higher level map. These characteristics reveal some tendency of a particular cortical area to be allocated for processing different types of neural signals that pertain to different classes of stimuli. Therefore, a cortical map with such organization gives rise to limited spam of representations. So, it is possibly a neural structure underlying the mental rigidity and repetitive behavior experienced by patients with ASD, as illustrated by Figure 6d.

**3.4.    Disruption of Long Distance Connections**

Our next experiments refer to alterations in connections between maps from distinct levels. These simulations thus involve the complete network, as pictured in Figure 1.

In particular, we are interested in impairments of axonal projections underlying the weakening of neural signals, and their disrupted propagation.

At first, we impose an alteration in the connection between *Prim1* and *Assoc* – connection 'A' in Figure 1. So, throughout the *Assoc* learning process, we set in 0.02 a low learning rate, $\rho$, for updating the weights relative to the projection departing from *Prim1*. With relation to the other projections in the network, the learning rates were kept in 0.5, as in the normal condition.

As we observe in Figures 7a-b, the damaged connection prevented the *Assoc* map from evolving to a neuron clusters-based organization. In this undeveloped second-level map, only a few isolated neurons are responsive to afferent signals (Figure 7a). Therefore, from such a reduced amount of neurons it will emerge what we previously called as *fixed signals*.

Consequently, the *Front* map presents an abnormal final state composed by two neuron groups, where three distinct kinds of stimuli are encoded by the same neurons in the larger cluster (Figure 7b). Thus, the fixed signals that emerged in *Assoc* gained an extremely large processing region throughout the *Front* map. Qualitatively, this situation is similar to what our results revealed in Section 3.2, Figure 6c. Here, however, the appearance of a markedly extensive area with overlapping codifications points to a more severe tendency of mental rigidity or repetitive behavior.

In the next experiment, we address the effect of a disruption between the *Assoc* and *Front* maps – connection 'B' in Figure 1. Such alteration influences only the *Front* map, whose final state is shown in Figure 8. Similarly to the modeling of the disruption in connection 'A', we set in 0.02 the learning rate, $\rho$, to update the weights relative to the projection 'B', during the *Front* learning process. The learning rates relative to the other projections in the network were kept in 0.5, as in the normal condition.

From Figure 8 we get that the disruption 'B' interferes with the *Front*'s development so that its final state presents three clusters – instead of the expected four-cluster organization – one of them with superposed codifications of two stimuli. This map organization indicates loss of selectivity, where a given group of neurons becomes responsive to stimuli from distinct class.

Now, we impose alterations in both 'A' and 'B' projections. Figure 9 shows the effect of such disruptions on the *Front* map formation. Here, we perceive a non consolidated map, where there is a tendency to the appearance of only two neurons groups, with overlapped codifications.

Overall, the simultaneous impairments in these two connections prevented the *Front* map from reaching the kind of pathological state observed in Figure 7b, where the 'A' projection alteration led to the emergence of an exacerbated area of superposed codifications – a plausible neural substrate of mental rigidity. On the other hand, when both 'A' and 'B' become weakened, the *Front* map's final state does not even reach a neural organization that allows a cognitive behavior oriented to mental rigidity or repetitive actions. In this map, concepts are not completely learned – as observed through the malformation of the clusters – and, thus, the cognitive deficits might be more severe due to some incapacity of topographically representing environmental information in the cerebral cortex.

### 3.5. Disruption of Long Distance Connections in presence of Excitatory Imbalances

Finally, we address the behavior of a network with disruptions in its long connections in presence of excitatory imbalances. Our simulations refer to the network architecture shown in Figure 1, with disruptions in connections 'A' and/or 'B', as modeled in Section 2.5. With relation to the excitatory imbalances, here our focus is on the over-strengthening of excitatory synapses. Thus we applied the modeling presented in Section 2.4, with $\Omega = 5$ in (2), where $\Omega$ is the exceeding amount due to the synaptic over-strengthening.

For each case of *Long Connection Disruption* addressed in Section 3.4 (disruption of 'A', disruption of 'B', or disruption of both 'A' and 'B'), we inflict excitatory imbalances in 11 different combinations of maps throughout the network, as: (*Prim1* and *Prim2*); (*Assoc*); (*Front*); (*Assoc* and *Front*); (*Prim1*, *Prim2* and *Assoc*); (*Prim1*, *Prim2* and *Front*); (*Prim1*, *Prim2*, *Assoc* and *Front*); (*Prim1*); (*Prim1* and *Assoc*); (*Prim1* and *Front*); (*Prim1*, *Assoc* and *Front*).

By bringing together such combinations of long distance disruptions and excitatory imbalances in different maps, we got a variety of outputs not observed in our previous simulations. Tables 2-4 present the descriptions of the final states achieved by each map in the network, as follows: Case 1 refers to the experiments concerning disruptions in connection 'A', Case 2 deals with disruptions in connection 'B', while Case 3 approaches disruptions in both connections 'A' and 'B'.

Considering, in each experiment, the organizational pattern in the highest level map, *Front*, with relation to the amount of clusters and presence of overlapping in neural codification, we observe that: in comparison to Cases 2 and 3, Case 1 presents a larger range of distinct abnormal final states – maps composed of 3, 2 or 1 cluster(s), 2 or 1 encoding neuron(s), with presence of overlapping in all situations. Whereas in Case 2 we note the presence of maps organized in (4, 3, 2 or 1) cluster(s), with overlapping in maps with 1, 2 and 3 clusters – with no final state composed by isolated encoding neurons. Similarly, Case 3 presents *Front* maps organized in (4, 3 or 1) cluster(s), with overlapping in maps with 1 and 3 clusters – again, with no final state composed by isolated encoding neurons.

At first sight, it seems strange that a network with only one disrupted connection – which is localized between its lowest levels maps - provides a greater amount of distinct abnormal final states than the other cases do. What this situation reveals, however, is the attenuating effects that emerge from the co-existence of long connection disruptions and repetitive, fixed neural signals produced by over-excited regions.

In fact, according to the dynamics of cortical maps, a steady and repetitive signal, over time, becomes strong enough to create a disproportionately large region to represent it in the following level's map of the network, as shown in the experiments presented in Sections 3.3-4. On the other hand, if such repetitive signals are conveyed to an over-excited region, the overall effect might be quite different, as we observe in Case 1 ('A' connection disrupted), when both *Assoc* and *Front* are under excitatory imbalance (4th line of Table 2). In this situation, the fixed signal produced by the over-excited *Assoc* does not give rise to an enlarged, one-cluster-based *Front* map. Instead, we have a *Front* map composed by a unique encoding neuron. Here, the influence of the excitatory imbalance in *Front* exceeds the impact of repetitive input signals.

Conversely, in both Cases 2 and 3, when *Assoc* and *Front* are again under over-excitatory imbalance, the presence of a disrupted connection between such maps changed the effect of the over-excitation on the *Front* map, in comparison with the situation just described in Case1. We observe in the fourth lines of Tables 3 and 4 that the fixed signals sent by the over-excited *Assoc* had to pass through a disrupted connection 'B', which attenuated the character of the signal being transmitted, so that it did not behave anymore as a strong, unique fixed signal. As a consequence, the signals departing from *Assoc* stimulated a broad area in *Front*, which was not neutralized by the over-excitatory effect on *Front*. Therefore, the final state of highest maps in the network is characterized by a cluster of neurons, and not by a unique, isolated encoding neuron.

| | **Maps under Excitatory Imbalance** | **Primary Maps Final State** | ***Assoc* Map Final State** | ***Front* Map Final State** |
|---|---|---|---|---|
| **Disruption in Connections 'A'** | *Prim1* and *Prim2* | 3 encoding neurons: 1 neuron encodes 2 classes of stimuli (overlapping) | Absence of clusters, Overlapping, No patterns | 3 clusters: 1 cluster presents 2 class-overlapping, takes approximately half of the map extension (195 encoding neurons) |
| | *Assoc* | Well-organized in 4 clusters | 1 encoding neuron: 4 class-overlapping | 1 cluster (242 encoding neurons): 4 class-overlapping |
| | *Front* | Well-organized in 4 clusters | 6 encoding neurons: 3 close neurons encode a given class, 3 neurons set apart from the others | 2 encoding neurons: 1 neuron encodes 3 classes of stimuli (overlapping) |
| | *Assoc* and *Front* | Well-organized in 4 clusters | 1 encoding neuron: 4 class-overlapping | 1 encoding neuron: 4 class-overlapping |
| | *Prim1*, *Prim2* and *Assoc* | 3 encoding neurons: 1 neuron encodes 2 classes of stimuli (overlapping) | 1 encoding neuron: 4 class-overlapping | 1 cluster & Isolated neurons: 4 class-overlapping |
| | *Prim1*, *Prim2* and *Front* | 3 encoding neurons: 1 neuron encodes 2 classes of stimuli (overlapping) | Absence of clusters, Isolated neurons: Overlapping, No patterns | 2 encoding neurons: 1 neuron encodes 3 classes of stimuli (overlapping) |
| | *Prim1*, *Prim2*, *Assoc* and *Front* | 3 encoding neurons: 1 neuron encodes 2 classes of stimuli (overlapping) | 1 encoding neuron: 4 class-overlapping | 1 encoding neuron: 4 class-overlapping |
| | *Prim1* | (*Prim1*) 1 encoding neuron: 4 class-overlapping, (Prim2) Well-organized in 4 clusters | 7 rather organized encoding neurons: Absence of clusters, No overlapping | 2 clusters: 1 cluster presents 3 class-overlapping, takes almost half of the map extension (184 encoding neurons) |
| | *Prim1* and *Assoc* | (*Prim1*) 1 encoding neuron: 4 class-overlapping, (Prim2) Well-organized in 4 clusters | 1 encoding neuron: 4 class-overlapping | 1 cluster (170 encoding neurons): 4 class-overlapping |
| | *Prim1* and *Front* | (*Prim1*) 1 encoding neuron: 4 class-overlapping, (Prim2) Well-organized in 4 clusters | 7 rather organized encoding neurons: Absence of clusters, No overlapping | 1 encoding neuron: 4 class-overlapping |
| | *Prim1*, *Assoc* and *Front* | (*Prim1*) 1 encoding neuron: 4 class-overlapping, (Prim2) Well-organized in 4 clusters | 1 encoding neuron: 4 class-overlapping | 1 encoding neuron: 4 class-overlapping |

**Table 2. Case1:** Descriptions of *Prim1*, *Prim2*, *Assoc* and *Front* maps final states in simulations involving both the disruption in connection 'A' and excitatory imbalances throughout the modeled neural network.

Another pattern of final state that is observed a number of times in *Front* is worth to be highlighted: the emergence of maps composed of four well-defined clusters. Since, in the scope of our work, this pattern reflects the behavior of a normal neural development, it is surprising that networks presenting several pathological alterations give rise to such well formed maps.

| | **Maps under Excitatory Imbalance** | **Primary Maps Final State** | ***Assoc* Map Final State** | ***Front* Map Final State** |
|---|---|---|---|---|
| **Disruption in Connection 'B'** | ***Prim1* and *Prim2*** | 3 encoding neurons: 1 neuron encodes 2 classes of stimuli (overlapping) | 2 clusters: 1 cluster presents 3 class-overlapping (125-47/47/47 encoding neurons) | Tendency to emergence of 4 shrunk clusters Isolated neurons |
| | *Assoc* | Well-organized in 4 clusters | 2 encoding neurons: 2 class-overlapping in each one | Tendency to emergence of 2 clusters[*]: 4 class-overlapping Isolated neurons |
| | *Front* | Well-organized in 4 clusters | Well-organized in 4 clusters | Well-organized in 4 clusters |
| | *Assoc* and *Front* | Well-organized in 4 clusters | 1 encoding neuron: 4 class-overlapping | 1 cluster & Isolated neurons: 4 class-overlapping |
| | *Prim1*, *Prim2* and *Assoc* | 3 encoding neurons: 1 neuron encodes 2 classes of stimuli (overlapping) | 1 encoding neuron: 4 class-overlapping | 1 cluster & Isolated neurons: 4 class-overlapping |
| | *Prim1*, *Prim2* and *Front* | 3 encoding neurons: 1 neuron encodes 2 classes of stimuli (overlapping) | 2 adjoining clusters: 3 class-overlapping in the smallest one | Well-organized in 4 clusters |
| | *Prim1*, *Prim2*, *Assoc* and *Front* | 3 encoding neurons: 1 neuron encodes 2 classes of stimuli (overlapping) | 1 encoding neuron: 4 class-overlapping | 1 cluster & Isolated neurons: 4 class-overlapping |
| | *Prim1* | (*Prim1*) 1 encoding neuron[*]: 4 class-overlapping, (Prim2) Well-organized in 4 clusters | Well-organized in 4 clusters | 3 shrunk clusters: 2 class-overlapping in one of them Isolated neurons separated in classes |
| | *Prim1* and *Assoc* | (*Prim1*) 1 encoding neuron: 4 class-overlapping, (Prim2) Well-organized in 4 clusters | 3 encoding neurons: 1 neuron encodes 2 classes of stimuli (overlapping) | Tendency to emergence of 2 clusters: 3 class-overlapping in one of them Isolated neurons |
| | *Prim1* and *Front* | (*Prim1*) 1 encoding neuron: 4 class-overlapping, (Prim2) Well-organized in 4 clusters | Well-organized in 4 clusters | 4 clusters & Few isolated neurons |
| | *Prim1*, *Assoc* and *Front* | (*Prim1*) 1 encoding neuron: 4 class-overlapping, (Prim2) Well-organized in 4 clusters | 3 encoding neurons: 1 neuron encodes 2 classes of stimuli (overlapping) | 2 clusters[*]: 1 exacerbate cluster presents 3 class-overlapping (154 encoding neurons) Isolated neurons |

**Table 3. Case 2:** Descriptions of *Prim1*, *Prim2*, *Assoc* and *Front* maps final states in simulations involving both the disruption in connection 'B' and excitatory imbalances throughout the modeled neural network.

We can see, however, through Table 3 (lines 3 and 6) and Table 4 (lines 6 and 10), the appearance of *Front* maps that are well-organized in four clusters. More, in Table 3 (line 10) and Table 4 (line 3), we note *Front* maps that, if not completely well-organized, present four clusters plus a few isolated neurons. Important to stress, the context that leads to the consolidation of these well or almost well-organized maps is not the same in Case 2 and in Case 3.

Such experiments show, again and more markedly, that attenuating effects come out due to the interaction between long connection disruptions and over-excitatory imbalances. In such experiments, there appear positive outcomes in the sense that those specific and simultaneous pathological alterations allowed the development of neural representations relative to normal cognition, according to our work frame.

|  | **Maps under Excitatory Imbalance** | *Prim*ary Maps Final State | *Assoc* Map Final State | *Front* Map Final State |
|---|---|---|---|---|
| **Disruptions in Connections 'A' and 'B'** | *Prim1* and *Prim2* | 3 encoding neurons: 1 neuron encodes 2 classes of stimuli (overlapping) | Sparce encoding neurons: Absence of clusters | Tendency to emergence of 4 clusters |
| | *Assoc* | Well-organized in 4 clusters | 1 encoding neuron: 4 class-overlapping | 1 cluster* & Isolated neurons: 4 class-overlapping |
| | *Front* | Well-organized in 4 clusters | Sparce encoding neurons: Absence of clusters | 4 clusters* & Few isolated neurons |
| | *Assoc* and *Front* | Well-organized in 4 clusters | 1 encoding neuron: 4 class-overlapping | 1 cluster & Isolated neurons: 4 class-overlapping |
| | *Prim1*, *Prim2* and *Assoc* | 3 encoding neurons: 1 neuron encodes 2 classes of stimuli (overlapping) | 1 encoding neuron: 4 class-overlapping | 1 cluster & Isolated neurons: 4 class-overlapping |
| | *Prim1*, *Prim2* and *Front* | 3 encoding neurons: 1 neuron encodes 2 classes of stimuli (overlapping) | Patternless map, 4 class-overlapping | Well-organized in 4 clusters |
| | *Prim1*, *Prim2*, *Assoc* and *Front* | 3 encoding neurons: 1 neuron encodes 2 classes of stimuli (overlapping) | 1 encoding neuron: 4 class-overlapping | 1 cluster & Isolated neurons: 4 class-overlapping |
| | *Prim1* | (*Prim1*) 1 encoding neuron: 4 class-overlapping, (*Prim2*) Well-organized in 4 clusters | 9 encoding neurons separated in classes: Absence of clusters | 3 shrunk clusters: 2 class-overlapping in one of them Isolated neurons separated in classes |
| | *Prim1* and *Assoc* | (*Prim1*) 1 encoding neuron: 4 class-overlapping, (*Prim2*) Well-organized in 4 clusters | 3 encoding neurons*: 1 neuron encodes 2 classes of stimuli (overlapping) | Tendency to emergence of 3 clusters*, 2 class-overlapping in one of them |
| | *Prim1* and *Front* | (*Prim1*) 1 encoding neuron: 4 class-overlapping, (*Prim2*) Well-organized in 4 clusters | Sparse encoding neurons separated in classes | Well-organized in 4 clusters |
| | *Prim1*, *Assoc* and *Front* | (*Prim1*) 1 encoding neuron: 4 class-overlapping, (*Prim2*) Well-organized in 4 clusters | 3 encoding neurons: 1 neuron encodes 2 classes of stimuli (overlapping) | 3 clusters*: 2 class-overlapping in one of them |

**Table 4. Case 3:** Descriptions of *Prim1*, *Prim2*, *Assoc* and *Front* maps final states in simulations involving both the disruptions in connections 'A' and 'B', and excitatory imbalances throughout the modeled neural network.

In the sequence, through Tables 5 and 6, we access the distribution of the different final states achieved by the *Assoc* and *Front* maps, in each addressed case. We can note that, in Case 1, where the disruption is restricted to connection 'A', final states composed of one encoding neuron predominate throughout the 2nd level maps. More, there is no *Assoc* map organized in clusters. On the other hand, in Case 2, where the disruption is restricted to connection 'B', although the majority of the final states in *Assoc* are maps composed of a few isolated encoding neurons, all the other maps are organized in clusters. With relation to Case 3, with disruptions in both 'A' and 'B', again, most of *Assoc* maps are composed of a few isolated encoding neurons - similarly to Case 2. But, the others maps left do not present clusters - here, we observe a significant amount of non-organized maps - similarly to Case 1.

Important to highlight, this similarity of patterns between the *Assoc* maps presented in Cases 1 and 3 does not lead to similar patterns between the corresponding *Front* maps in these cases. As we observe in Table 6, with relation to Case 1, in 5 out of 11 experiments, the *Front*'s final states are maps composed of one or two encoding neurons. While, in Case 3, no *Front* map presents such patterns.

| | *Assoc* Maps Configurations | | | | | | | | | |
|---|---|---|---|---|---|---|---|---|---|---|
| | 4 clusters | Absence of clusters (>= 9 neurons) | 2 clusters | Tendency to 2 clusters | 9 organized encoding neurons | 7 organized encoding neurons | 6 encoding neurons | 3 encoding neurons | 2 encoding neurons | 1 encoding neuron |
| **Case1** | 0 | 2 | 0 | 0 | 0 | 2 | 1 | 0 | 0 | 6 |
| **Case2** | 3 | 0 | 1 | 1 | 0 | 0 | 0 | 2 | 1 | 3 |
| **Case3** | 0 | 4 | 0 | 0 | 1 | 0 | 0 | 2 | 0 | 4 |

**Table 5.** Final states achieved by *Assoc* maps in each addressed case.

Again, in Case 1, we have no *Front* maps organized in 4-clusters maps. In Case 3, however, 4 out of 11 experiments reveal *Front* maps that present around 4 clusters. The same amount is observed in Case 2.

| | *Front* Maps Configurations | | | | | | | | | |
|---|---|---|---|---|---|---|---|---|---|---|
| | Well-organized 4 clusters | 4 clusters & Isolated neurons | Tendency to 4 clusters | 3 clusters | Tendency to 3 clusters | 2 clusters | Tendency to 2 clusters | 1 cluster & Isolated neurons | 1 cluster | 2 encoding neurons | 1 encoding neuron |
| **Case1** | 0 | 0 | 0 | 1 | 0 | 1 | 0 | 2 | 2 | 1 | 4 |
| **Case2** | 2 | 1 | 1 | 1 | 0 | 1 | 2 | 3 | 0 | 0 | 0 |
| **Case3** | 2 | 1 | 1 | 2 | 1 | 0 | 0 | 4 | 0 | 0 | 0 |

**Table 6.** Final states achieved by *Front* maps in each addressed case.

Apart from these extreme situations, in both Cases 1 and 3, we perceive that 4 out of 11 *Front* maps are composed of 1 cluster (& isolated neurons). And, almost the same result is observed in Case 2, where 3 experiments provide *Front* maps organized in 1 cluster & isolated neurons.

Concerning around-3 cluster *Front* maps, both Cases 1 and 2 present only one map with such pattern, while in Case 3 we note that three *Front* maps have this structure. Also, there is no *Front* maps composed of around 2 clusters, while such pattern is observed once in Case 1 and three times in Case 2.

Now, we would like to call attention to the relationship between the most frequent kind of final states in the *Assoc* and *Front* configurations. According to Table 5, we note that the 1-encoding-neuron is the more often configuration among the *Assoc* maps. And, through Table 6, we observe that the 1-cluster (& isolated neurons) is the one that most appears throughout the *Front* maps configurations. In fact, a closer analysis of the maps achieved through the simulations in Cases 1, 2 and 3 - Tables 2-4 - exposes a link between the most frequent patterns in *Assoc* and *Front*.

With relation to the 1-encoding-neuron pattern in the *Assoc* maps, 13 out of 33 experiments present such final state: 6 in Case 1, 3 in Case 2 and 4 in Case 3. And, through Table 7 we get that, except for three experiments, this final state in *Assoc* is followed by the 1-cluster (& isolated neurons) pattern in *Front*. Therefore, our results point to a link between these kind of patterns in our modeled networks, which incorporate alterations observed in studies of autism.

|  | *Assoc* Map | *Front* Map | |
|---|---|---|---|
|  | 1 encoding neuron | 1 cluster (& Isolated neurons) | 1 cluster (& Isolated neurons) preceded by the 1 encoding neuron map at *Assoc* |
| **Case1** | 6 | 4 | 3 |
| **Case2** | 3 | 3 | 3 |
| **Case3** | 4 | 4 | 4 |

**Table 7.** Influence of 1-encoding-neuron *Assoc* maps on the development of 1-cluster *Front* maps.

Finally, we underline the presence of well organized, 4-cluster maps at the third level of networks, even in presence of excitatory and connectivity imbalances. According to Table 8 bellow, in both Cases 2 and 3, two *Front* maps achieve a configuration relative to normality. These situations stress the regulatory effect that different alterations exert upon each other throughout a network architecture.

|  | *Assoc* Map | | | *Front* Map | | | |
|---|---|---|---|---|---|---|---|
|  | Tendency to 2 clusters | Well-organized 4 clusters | Absence of clusters (more than 9 neurons) | Well-organized 4 clusters | Well-organized 4 clusters preceded by 4 cluster map | Well-organized 4 clusters preceded by around-2-cluster map | Well-organized 4 clusters preceded by map with no clusters |
| **Case1** | 0 | 0 | 0 | 0 | 0 | 0 | 0 |
| **Case2** | 1 | 3 | 0 | 2 | 1 | 1 | 0 |
| **Case3** | 0 | 0 | 2 | 2 | 0 | 0 | 2 |

**Table 8.** *Assoc* maps influence on the formation of well-organized 4-cluster *Front* maps.

We can note that, in Case 2, from the two well-organized *Front* maps, one of them is preceded by an also well-organized *Assoc* map, while the other is preceded by a map with two adjoining clusters. And, both 4-cluster *Front* maps, in Case 3, are preceded by *Assoc* maps with no clusters.

In the sequence, we propose relationships between the patterns achieved through the simulations of our modeled networks, and cognitive symptoms observed in autism.

### 3.6. Maps Organizational Patterns and Putative Autistic Symptoms

For a satisfactory comprehension of our approach, it is necessary to clarify how a given pattern of neural activation in a cortical map is related to a given cognitive characteristic. So we devote this Section to make this relationship as clear as possible.

One principle behind our modeling is that cortical maps topologically codify information coming from both external and internal stimuli. A specific group of neurons, thus, becomes responsive to a specific class of stimuli, as a result of processes that occur during the cortical development. Apparently simple, this concept presents deep consequences concerning cognition, due to the topological aspect of the cortical maps organization. Besides the specificity between neuron clusters and input stimuli, these neuron groups keep a topological relationship in the sense that the more similar the stimuli, the closer the neurons that codify them. Neurons from a given cluster share the same neighborhood and thus are responsive to stimuli pertaining to a given class of stimuli, which share some kind of similarity. More than organization, such principle propitiates cognitive features as selectivity and generalization. In addition, another property also rules the process that conducts the development of a cortical map: the more often an input is, the largest the neural representation it acquires in the map, so that a greater amount of neurons becomes responsive to such an input pattern. Therefore, its representativeness gains more importance, in comparison to the other stimuli's.

Going further with these properties, we have that a neuron cluster, in which its neurons codify overlapped patterns (pertaining to distinct classes), are neural structures that promote absence of selectivity in cognitive processes. More, a large and unique population of neurons that responds to all classes of stimuli may be indicative of a neural structure that contributes to mental rigidity and repetitive behavior, because no matter how different the information that reaches the map is, the neural answer will be the same.

Important to highlight, as we assume that cortical maps are the basic neural structures, from which cognitive features will emerge, our understanding establishes that cognition depends on sensory codification. Also, all these principles are gathered in our model, since SOM networks incorporate such neurodevelopment properties, as exposed in Sections 1 and 2.2.

As our simulations incorporate excitatory imbalances and connectivity disruptions reported in studies about autistic brains, and as the properties above mentioned happen thanks to a complex network of neurobiological events that are dependent on both neurochemical balances and adequate brain connections, it is natural that our results reveal a range of different, well defined patterns characterizing the maps final states throughout the simulations of our modeled network.

Taking a step ahead, we examined plausible cognitive consequences that could be associated to such different kinds of maps patterns, in the context of the ASD. So, as a result, this analysis indicated possible autistic cognitive symptoms related to the final states maps obtained in our simulations. These relationships are summarized in Table 9.

Although massively parallel and distributed, information processing throughout the brain is hierarchically organized. Distinct classes of sensory stimuli are processed through distinct levels of neural networks, so that the lowest levels are in charge of more concrete information that, as the processing reaches the highest levels, becomes more abstract. One of the reasons why it happens consists in the gradual integration – through successive processing levels - of distinct kinds of signals coming from distinct primary areas.

In our model, the *Front* maps represent cortical areas in charge of high level cortical processing, which is markedly involved in the control of executive functions in the cognitive apparatus. Therefore, without lack of generality, departing from our computational simulations, we hint plausible neural origins for some autistic cognitive alterations based on the *Front* maps' final states, which emerge as a consequence of the excitatory imbalances and connectivity alterations inflicted throughout the modeled neural network.

So, a well-organized map, without overlapping, where its clusters are balanced and reflect the inputs' characteristics, is associated to a *normal cognitive condition*, in particular because this neural structure promotes selectivity and generalization. On the other hand, a map that presents a short reduction in the amount of clusters - in comparison to the normal, well-organized one - and moderate overlapping, although keeping a balanced, clustered organization, can be a neural substrate of mild lack of selectivity (due to the overlapping), which may bring a *mild mental rigidity*: in this case, some distinct inputs evoke the same neural response. There is a diminishing in the variety of neural responses.

| Front Map Outputs | Putative Cognitive Symptoms |
|---|---|
| - Well organized map, with 4 clusters and no overlaps.<br><br>- 4 clusters and a few isolated neurons | **Normal Cognitive Condition** |
| 3 small and organized clusters, one of them is responsive to 2 classes of stimuli. | **Mild Cognitive Damage/Mild Mental Rigidity** |
| - 3 clusters, one of them is responsive to 2 classes of stimuli and is larger than the other 2 clusters.<br><br>- 2 clusters, one of them is responsive to 3 classes of stimuli and is much larger than the other cluster. | **Mental Rigidity** |
| - 1 cluster (with or without a few isolated neurons) responsive to 4 classes of stimuli.<br><br>- Tendency to formation of 2 clusters, with neurons responsive to 3 classes of stimuli *. | **Marked Mental Rigidity** |
| Tendency to formation of 3 or 4 clusters, with overlaps. | **Expressive Cognitive Damage** |
| - 1 encoding neuron responsive to 4 classes of stimuli.<br><br>- 2 encoding neurons, one of them is responsive to 3 classes of stimuli. | **Extreme Cognitive Damage** |

**Table 9.** Putative cognitive symptoms associated to different outputs observed in the *Front* maps.

If, however, some cluster acquires a cortical area that is much greater than each region occupied by the other neuron groups, it implies a situation where a markedly large area is allocated to process a given kind of input. As a consequence, such a class of inputs will gain more control over cortical mechanisms, more importance and more representativeness in the mental life. This structure can be seen as a neural substrate for the appearance of a *marked mental rigidity*. A more severe condition arises when such a markedly larger cluster presents overlapping - simultaneous codification of different classes of stimuli: in this case, distinct input classes evoke the same cortical response. Or, no matter what input reaches a high cortical area, the effect will be the same. Such neural structure, if does not give origin, at least contributes to the emergence of the autistic repetitive behavior.

Another kind of final state that our simulations revealed consists of maps that do not achieve a well defined organization in clusters. Although they present a tendency to the emergence of neuron groups, these are not actually formed. It points to an abnormal, poor representation of concepts, which possibly leads to *expressive cognitive damages*, due to the brain inability to codify and retain a satisfactory representation of the world. A natural consequence of such misrepresentation consists of a diffuse, weakened neural response, since the amount of neurons that should be involved in such command is not completely active.

At last, we observed an extreme situation where no pattern is established, because only a tiny amount of neurons codifies information. Therefore, the information that reaches this kind of map is not processed and, as a consequence, such structure is able to give rise to an *extreme cognitive damage* - no brain response to stimuli.

Now, we revisit pathological alterations imposed on the model, and assign putative autistic symptoms to some of these cases. Firstly, we reexamine the experiment referring to the influence that an excitatory imbalance in a low level map exerts on a higher level map with no disruption in the long connection between these maps. We observe, through Table 10, the main features observed in the maps' organizational patterns, which - according to our proposal - are associated with manifestations of *mental rigidity*.

| Excitatory Imbalance WITHOUT Long Connections Disruptions | |
|---|---|
| **Low Level** | Lack of encoding neurons. Absence of organization in clusters. Overlapping codifications. |
| **High Level** | Repetitive signals. Overlapping codifications. Reduction in the amount of clusters of neurons. |
| **Mental Rigidity** | |

**Table 10.** *Mental rigidity*, the putative symptom associated to the main features observed in the maps developed under excitatory imbalance.

| Long Connections Disruptions WITHOUT Local Excitatory Imbalance | | |
|---|---|---|
| **1st to 2nd Level** | **2nd to 3rd Level** | **1st to 2nd and 2nd to 3rd Level** |
| Lack of encoding neurons at the 2nd level. Predominance of repetitive signals. Overlapping codifications in an extremely enlarged cluster of neurons at the 3rd level. Loss of selectivity. | Reduction in the amount of clusters of neurons at the 3rd level. Incipient organization in small clusters at the 3rd level. Overlapping codifications. Loss of selectivity. | Absence of organization in clusters at the 3rd level. Overlapping codifications in a rather formed cluster at the 3rd level. No selectivity. |
| **Marked Mental Rigidity** | **Mild Cognitive Damage / Mild Mental Rigidity** | **Expressive Cognitive Damage** |

**Table 11.** Putative cognitive symptoms associated to the main features observed in the maps developed under long connections disruptions.

In the sequence, Table 11 shows the maps' characteristics and putative symptoms related to them according to the experiments involving long connections disruptions, but no local excitatory imbalance. As we observe, a *marked mental rigidity* would come from connectivity disruptions between the 1st and 2nd levels, while a *mild mental rigidity* would arise from disruptions in the connections between the 2nd and 3rd levels. The simultaneous occurrence of disruptions between the three levels, however, would produce *expressive cognitive damage*, as defined above.

Finally, we merged the results regarding our experiments on long connections disruptions in presence of local excitatory imbalances - available in Tables 2-4 - with those proposed relationships between *Front* maps final states and autistic putative cognitive symptoms - presented in Table 9. So, through Table 12, we can find the putative symptom associated to a specific combination of connectivity disruption and local excitatory imbalance, which were addressed by our simulations. For instance, in our model, if the connection between *Prim1* and *Assoc* (1st and 2nd levels) is disrupted, and the *Prim1*, *Prim2* and *Assoc* maps are under excitatory imbalances (3rd level not over-excited), the *Front* map final state presents characteristics that are plausible neural substrates of *marked mental rigidity*. Through Figures 10a,b we provide a pictorial representation of the ideas presented in this section.

| Long Connections Disruptions **WITH** Excitatory Imbalance ||||||
|---|---|---|---|---|---|
| 1st to 2nd Level || 2nd to 3rd Level || 1st to 2nd and 2nd to 3rd Levels ||
| 3rd Level NOT Over-Excited | 2nd Level NOT Over-Excited | 1st Level Over-Excited (only) | *Prim1* and *Prim2* Over-Excited | 1st Level completely Over-Excited (only) | Expressive Cognitive Damage |
| | Mental Rigidity | | Expressive Cognitive Damage | | |
| | | | *Prim1* Over-Excited (only) | 2nd Level Over-Excited | *Prim1* (and NOT *Prim2*) Over-Excited |
| | | | Mild Cognitive Damage | | Mental Rigidity |
| | 2nd Level Over-Excited | 2nd Level NOT Over-Excited | 3rd Level Over-Excited | | Otherwise |
| | Marked Mental Rigidity | | Normal Cognition | | Marked Mental Rigidity |
| 3rd Level Over-Excited | Extreme Cognitive Damage | 2nd Level Over-Excited | Marked Mental Rigidity | 3rd Level Over-Excited | Normal Cognition |

**Table 12.** Putative cognitive symptoms associated to the main features observed in the maps developed under excitatory imbalance and long connections disruptions.

## 4. Discussion

The mysterious autistic behavior seems, in many cases, inaccessible. However, the inclusion of ASD patients in schools and work activities started revealing possibilities and achievements unknown before, when they were commonly set apart from the social life (Grinker, 2007), (Camargo and Bosa, 2009), (Della Barba and Minatel, 2013). Indeed, due to the brain plasticity mechanisms, stimuli enriched environments generally are much better for the neurodevelopment than poor ones.

On the other hand, the brain structural impairments observed in the ASD interfere in the consolidation and functioning of crucial neural circuits, and still impose serious limitations to the adaptation of ASD patients - and their families - to daily routines.

Throughout this work we look at some excitatory and connectivity alterations previously evidenced in autistic brains. Thus, here, we deal with factors that prevent neural systems from reaching a satisfactory development and functioning in ASD patients.

Based on both clinical and experimental evidences, we not only employed Kohonen's SOM algorithm to represent a brain mechanism that underlies the ability of forming categories. More, we addressed connectional features, via a three-multilayer SOM network, representing distinct levels of cortical processing, and simulated a number of *in silico* experiments. From our results, we infer a way by which the addressed connectivity disruptions give rise to some cognitive difficulties observed in the ASD, as lack of categorization, selectivity and generalization abilities, and discuss their relationship with mental rigidity and repetitive behavior.

In fact, the capacity of constructing conceptual classes and selectively deciding to which one a given element pertains (Minshew *et al*., 2002), is in the core of the reasoning through categories. Therefore, it underlies the ability of making generalizations, which is of essential importance for the emergence of the abstract reasoning. For instance, when someone is not able of perceiving that, although different, objects may share features that impose some degree of similarity between them, it turns to be difficult the use of abstract constructions, as words and languages, or even making simple inferences concerning equalities and belongings. Poor categorization mechanisms can thus bring together a tendency to maintain invariant habits and ideas, in particular when similar situations are considered as different ones, to which apparently there is no behavioral reference.

From a neurobiological point of view, cortical maps – with their topologically organized neuron clusters – can be plausibly considered as fundamental components of the neural substrates for selectivity and generalization. So, by computationally simulating the development of a network composed of maps, with imposed alterations reflecting ASD impairments previously evidenced in neuroscience studies, it was possible - at least theoretically - to associate the neural basis of categorical reasoning to some pathological mechanisms in the autistic cognition.

Our first set of simulations, presented in section 3.2, explores the central idea exposed in Rubenstein and Merzenich (2003). They propose that an excess of cortical excitatory level underlies many autistic symptoms, since it could make the cortex more excitable and less functionally differentiated. Such high excitatory state would be due to an imbalance between excitatory and inhibitory synapses, in which case there would be much more excitation than in the non-autistic brain.

By over-strengthening the excitatory synapses in a single map development, we verified two kinds of situations. In the first one, the map did not reach a stable final state. Or, neuron groups do not achieve a final configuration, due to the continuous changes. Therefore, these results suggest that categories remain undefined, with no concepts consolidation. In the second situation, however, where the excitatory synapses are even stronger, neurons do not organize in clusters: only a few of them respond to neural signals, and become codifiers of stimuli. So, no matter its kind, information ie very poorly - if not at all - represented in the brain. In this aberrant case, there is no neural structure to subserve categorization mechanisms, and stimuli tend to be not differentially codified by a restrict number of neurons.

Again through single maps simulations, we looked at excitatory imbalances under a different approach: by enlarging the excitatory influence area of each neuron. Here, we observed that, for values above a defined normal range, larger excitatory areas – throughout the map development – lead to smaller clusters of encoder neurons. Eventually, the largest excitatory area simulated gave risen to an underdeveloped map, similar to the above mentioned case. Overall, this set of experiments indicates that, in some cases, markedly high local excitatory states originate neural structures - in this case, neuron clusters - relative to smaller classes that are not able to encode the richness of the environmental stimuli. Then, we suspect that this kind of cortical over-excitation is linked to the representation of poor and difficult to enhance associations classes, and also to less selectivity - due to encoding superposition.

In the sequence, in Section 3.3, we use a two-level network to address the situation where a degenerate map sends efferent projections to a higher level network area. More specifically, we investigate how the local excitatory over-strengthening of synapses in lower level maps influences the self-organization of an upper area. This experiment reveals a higher level map with characteristics incompatible with selectivity, i.e. superposed neural codifications. Besides that, there is a very large area dedicated to a unique signal type, which indicates an exaggerated neural allocation for a restrict representation or functionality. It is plausible to suggest, thus, that such a pathological structure underlies both mental rigidity and repetitive behavior in ASD: since this pathological representation pervades an anomalous huge area, it dominates the individual's mental life and, consequently, his/her actions.

Interestingly, according to our results, the over excitatory synaptic state during the development of the lower level map caused the appearance of neural substrates for mental rigidity and lack of selectiveness only in the posterior network level. It points to the global, network-based, character of the cognitive processing in the brain, where a given behavioral symptom can come from distinct kinds of alterations, in distinct cortical areas.

In addition, the highlighted cluster enlargement can be associated to the noisy cortical excitability and the lower differentiation proposed by Rubenstein and Merzenich (2003), who suggested that such pathological excitability would cause the strong aversive reaction observed in autistic children, especially in presence of noisy environments. Here, it is important to note that, accordingly to our simulations, the excitatory unbalance presents different manifestations, depending on the stage of an individual's life, i.e., during the neurodevelopment, when the organization of the cortical maps takes place, high excitatory states would prevent the emergence of a neural topography that would be responsible for a structured perception of the world.

Disrupted connectivity is the approach we undertook in the following simulations. Numerous studies point to complex anatomical abnormalities in long connections throughout brains of ASD patients–(Courchesne and Pierce, 2005), (Geschwind and Levitt, 2007), (Zikopoulos and Barbas, 2010), (Kana *et al*., 2011), (Mostofsky and Ewen, 2011), (Schipul *et al*., 2011). Indeed, the diffusion tensor imaging (DTI) technique has been revealing alterations in white matter structural integrity, i.e. axonal properties variations due to their thickness alterations, and reduction of long distance pathways. Overall, such impairments hinder the transmission of nervous impulses, making the neural signaling weaker, slower, unsynchronized or even absent.

In Section 3.4, our experiments illustrate the weakening of projections between maps from different levels throughout a three-level network. In the first simulation, the enfeeblement only occurs in the connection between a first level map and the second, intermediary, level. The effects, however, are present in both second and third level maps: in the *Assoc* map there are only a few encoding neurons that are not organized in clusters, while, in the *Front* map, occurs codification overlapping along markedly enlarged neuron clusters. This experiment indicates that a connectivity impairment, which weakens the neural signal, can distinctively alter the consolidation of maps in quite distant cortical regions, in particular, the maps responsible for the higher order executive functions. They also point to damages that mar functionalities associated to biological mechanisms for categorization.

In general, these results are consistent with ASD studies pointing to disconnection of regions involved in higher-order associations (Geschwind and Levitt, 2007), and under-connectivity of long distance cortical connections, manifested in complex and social functions (Kana *et al*., 2011). More particularly, some characteristics underlying the *Front* map, – as lower number of clusters, if compared with normal condition, besides a very large area where neurons respond for stimuli from distinct classes – agree with and illustrate the idea exposed by Courchesne and Pierce (2005). According to these authors, connectivity within frontal lobe is disorganized and poorly selective as long as under-connectivity between frontal cortex and other regions impairs the frontal integrative function, promotes lack of synchronization, and degenerates information representation. In addition, the converse situation, where a too large area in the *Front* map processes only a particular kind of signal, hints to a pathological over-excitatory process possibly associated to the hypothesis that prefrontal areas are locally over-connected in the ASD (Zikopoulos and Barbas, 2010).

Going further with the matter, our simulations are in line with evidences pointing to relationships between long distances under-connectivity and local frontal high excitability and, thus, over-connectivity, in ASD (Zikopoulos and Barbas, 2010), (Kana *et al*., 2011). Therefore, we corroborate the unifying approach that considers such disorder as composed by connectional impairments relative to both over and under-connectivity (Courchesne and Pierce, 2005), (Kana *et al*., 2011).

In the next experiment, also in Section 3.4, we created another connection disruption in the network, between the second and the third levels maps. As a result, the *Front* map – representative of a frontal area – presents a non organized final state, with coding superposition throughout a unique cluster rather formed: an isolated neural structure that prevents selectivity, and triggers undifferentiated responses from an exacerbated number of neurons – as if, no matter what signals stimulate it, a large area is always allocated to produce similar responses.

In Section 3.5, we combine the effects of excitatory synapses over-strengthening and axonal disruption. The conjunction of these two factors propitiated the appearance of many degrees of differentiation between the patterns observed in each case separately. Such organizational patterns variety ranges from extremely damaged maps with almost any responsive neurons up to completely well-formed maps, which arise from the annulment actions that those factors can infringe upon each other.

Throughout Section 3.6, we explicit the relationships between the types of patterns provided by our simulations and cognitive features observed in ASD. Since, ultimately, cortical maps are neural basis for concepts representation, their organization points to the way the brain might be dealing with processes underlying categorization, generalization and abstraction.

Overall, our simulations suggest that both excitatory synapses over-strengthening and disruption of long distance projections are able to generate neural structures that contribute to the appearance of mental rigidity and repetitive behavior, because they do not present the adequate topographic organization responsible for selectivity, categorization and neural responses the reflect the types and frequencies of the stimuli to whom an individual is exposed. According to our experiments, excitatory synapses over-strengthening diminishes the amount of encoding neurons in a map, which leads to the repetitive propagation of similar neural signals, and the consequent appearance of enlarged areas in posterior levels maps. On the other hand, the disruption of long distance projections tends to equalize the differences between neural signals. Therefore, it generates a great repetition of very similar inputs into a map, which eventually lead to the formation of enlarged clusters of neurons.

Although out of this work scope, our model incorporates a parameter associated with dopamine action in the modulation of the encoding neurons behavior. It is the threshold $\Theta$, which defines the minimal percent of the *winner* activation that a neuron must achieve to become an encoding neuron (Carvalho *et al*., 2003), (Mendes *et al*., 2004). Since clinical and experimental evidences indicate abnormal dopaminergic activity in patients with ASD (Ernst *et al*., 1997), (Narita *et al*., 2002), (Muhle *et al*., 2004), (Mittleman *et al*., 2008), (Nakasato *et al*., 2008), (Nguyen *et al*., 2014), (Paval, 2017) and dopamine seems to participate in the control of signal to noise ratio in the cortex (Servan-Schreiber *et al*., 1990), (Durstewitz *et al*., 2000), (Rubenstein and Merzenich, 2003), such parameter can be more explored in future models addressing dopaminergic imbalances in ASD. See also (Kriete and Noelle, 2015) for a model of dopaminergic action in the ASD. Important to note, the model proposed by Guimarães (2018) indicates that dopaminergic imbalances in subcortical structures of the reward circuit also contribute to the appearance of repetitive signals and mental rigidity in ASD patients.

At last, we would like to note that a bottom-up model allows for a very specific approach - the way neural signals propagate from low processing levels up to higher ones. Keeping in mind, however, that cortical communication is much more complex, thus involving top-down and massively parallel pathways, further works should account for bi-directional connections between frontal and other brain areas. It would convey a way of investigating the influence of pathological higher cortical areas on the functioning of primary ones.

**5.     Conclusions**

Using computational simulations to address the neural mechanisms of competition and cooperation, we investigated local excitatory imbalances – particularly, the excess of excitatory synaptic influence that spoils the cooperation step throughout the development of cortical maps. Since we assume that cortical maps are fundamental parts of the neural substrate for categorization, our first results indicate that such imbalances reduce both categorization and generalization abilities in ASD, as illustrated in the Figure 11. The simulations relative to the weakening of long connections concern the lack of neural integration. They evidence a poorly differentiating processing in higher cortical areas, which possibly underlies the disruption of selectivity mechanisms, besides the appearance of neural structures plausibly compatible with autistic characteristics as topics preferences and resistance to changes.

Our results indicate that, in addition to the situations where only a local alteration occurs, a particular autistic cognitive deficit could be caused also by impairments throughout distinct brain areas simultaneously. Or, it could even be the consequence of some region malfunction due to structural impairments present in distant sites. So, the richness of symptoms in the ASD would reflect the systemic cerebral behavior.

It is important to note that although our model is a plausible representation of the mind, all of our theoretical results need to be experimentally validated in the future.

Summarizing, in this work, we presented a theoretical model that explains the appearance of specific cognitive impairments clinically observed in ASD, in terms of excitatory unbalances and connectivity alterations described in neuroscience studies of such disorder.

**Acknowledgements:** We gratefully thank Alexandre L. Madureira for encouraging and coordinating this research, the Brazilian agency CNPq (grant: PDI 560108/2010-9C) for the financial support, Frédéric Valentin for the support and hospitality, and John R. Kenower for his careful review of the manuscript. We also recognize the ADACA Lab as a stimulating environment for studies concerning autistic cognition. This work is dedicated to João H. M. Nascimento, and his mother Eloísa.

**References**

American Psychiatric Association. (2013) Diagnostic and statistical manual of mental disorders: DSM-5. Washington, D.C: American Psychiatric Association.

Bosl, W., Tierney, A., Tager-Flusberg, H., Nelson, C. (2011) EEG complexity as a biomarker for autism spectrum disorder risk. *BMC Medicine*, 9: 18, doi:10.1186/1741-7015-9-18.

Boucher, J. (2012) Research review: Structural language in autistic spectrum disorder – characteristics and causes. *The Journal of Child Psychology and Psychiatry*, 53(3): 219-233.

Buonomano, D.V., Merzenich, M.M. (1998) Cortical plasticity: From synapses to maps. *Annual Review of Neuroscience*, 21: 149-186.

Camargo, S.P.H., Bosa, C.A. (2009) Competência social, inclusão escolar e autismo: Revisão crítica da literatura. Psicologia & Sociedade, 21 (1): 65-74.

Caminha, V.L.P.S., Caminha, A.O.,Vicente, G.L.F., Assis, L.M., Huguenin, J.Y., Alves, P.P., Felix, P.C., Pimentel, R.D.P. (2013) Ambiente Digital de Aprendizagem para Crianças Autistas (ADACA) Anais do II Fórum Internacional de Inclusão: Discutindo Autismo e Deficiência Múltipla, 1:156 -167.

Carvalho, L.A.V., Ferreira, N.C., Fiszman, A. (2001) A theoretical model for autism. *Journal of Theoretical Medicine*, 3: 271–286.

Carvalho, L.A.V., Mendes, D.Q., Wedemann, R.S. (2003) Creativity and delusions: The dopaminergic modulation of cortical maps. *Lecture Notes in Computer Science*, 2657: 511-520.

Cohen, I. L. (1994) An artificial neural network analogue of learning in autism. *Biological Psychiatry*, 36: 5–20.

Courchesne, E., Pierce, K. (2005) Why the frontal cortex in autism might be talking only to itself: Local over-connectivity but long-distance disconnection. *Current Opinion in Neurobiology*, 15:225-230.

Della Barba, P.C. S., Minatel, M.M. (2013) Contribuições da terapia ocupacional para a inclusão escolar de crianças com autismo. *Cadernos de Terapia Ocupacional da UFSCar*, 21(3): 601-608.

Durstewitz, D., Seamans, J.K., Sejnowski, T.J. (2000) Dopamine-mediated stabilization of delay-period activity in a network model of prefrontal cortex. *Journal of Neurophysiology*, 83: 1733-50.

Ernst, M., Zametkin, A.J., Matochik, J.A., Pascualvaca, D., Cohen, R.M. (1997) Low medial prefrontal dopaminergic activity in autistic children. *The Lancet*, 350(9078): 638.

Fernandez, E. A., Balzarini, M. (2007) Improving cluster visualization in self-organizing maps: Application in gene expression data analysis. *Computers in Biology and Medicine*, 37(12):1677–1689.

Flippin, M., Reska, S., Watson, L.R. (2010) Effectiveness of the Picture Exchange Communication System (PECS) on communication and speech for children with autism spectrum disorders: A meta-analysis. *American Journal of Speech-Language Pathology*, 19(2): 178-195.

Frost, L. A., Bondy, A. (1994). PECS: The Picture Exchange Communication System. Cherry Hill, NJ: Pyramid Educational Consultants.

Geschwind, D. H. (2009) Advances in autism. *Annual Review in Medicine*, 60: 367-380.

Geschwind, D. H., Levitt, P. (2007) Autism spectrum disorders: Developmental disconnection. *Current Opinion in Neurobiology*, 17: 103-111.

Gogolla, N., LeBlanc, J.J., Quast, K.B., Südhof, T.C., Fagiolini, M., Hensch, T.K. (2009) Common circuit defect of excitatory-inhibitory balance in mouse models of autism. *Journal of Neurodevelopmental Disorders*, 1: 172-181.

Goldstein, G., Allen, D., Minshew, N.J., Williams, D.L., Volkmar, F., Klin, A., Schultz, R. (2008) The structure of intelligence in children and adults with high functioning autism. *Neuropsychology*, 22(3): 301-312.

Grinker, R.R. (2010) Autismo: Um mundo obscuro e conturbado. Trad. Catharina Pinheiro, São Paulo: Larousse do Brasil.

Guimarães, Karine. (2018) Extension of Reward-Attention Circuit Model: Alcohol's Influence on Attentional Focus and Consequences on Autism Spectrum Disorder. *Neurocomputing,* 10.1016/j.neucom.2018.10.034.

Gustafsson, L. (1997) Changes Inadequate cortical feature maps: A neural circuit theory of autism. *Biological Psychiatry*, 42: 1138-1147.

Happe, F., Frith, U. (2006) The Weak Coherence Account: Detail-focused Cognitive Style in Autism Spectrum Disorders. *Journal of Autism and Developmental Disorders*, 36(1): 5-25.

Haykin, S. (1999) Neural Networks: A Comprehensive Foundation, 2nd Edition, Prentice-Hall.

Hill, E.L., Frith, U. (2003) Understanding autism: insights from mind and brain. *Philosophical Transactions of the Royal Society of London B*, 358: 281-289.

Iacoboni, M. (2009) Imitation, empathy, and mirror neurons. *The Annual Review of Psychology*, 60: 653-70.

Johnston, S. P. (2015) A cultural comparison of applied behavior analysis for autism spectrum disorder. Scripps Senior Theses. Paper 540.

Kana, R.K., Libero, L.E., Moore, M.S. (2011) Disrupted cortical connectivity theory as an explanatory model for autism spectrum disorders. *Physics of Life Reviews*, 8: 410–437.

Klin, A., Jones, W., Schultz, R., Volkmar, F. (2003) The enactive mind, or from actions to cognition: lessons from autism. *Philosophical Transactions of The Royal Society of London B*, 358: 345-360.

Kohonen, T. (1982) Self-organized formation of topologically correct feature maps. *Biological Cybernetics*, 43: 59-69.

Kriete, T., Noelle, D.C. (2015) Dopamine and the Development of Executive Dysfunction in Autism Spectrum Disorders. *PLoS ONE*, 10(3): e0121605. doi:10.1371/journal.pone.0121605.

Lazarev, V.V., Pontes, A., Mitrofanov, A.A., de Azevedo, L.C. (2010) Interhemispheric asymmetry in EEG photic driving coherence in childhood autism. *Clinical Neurophysiology*, 121(2): 145-152.

Lazarev, V.V., Pontes, A., Mitrofanov, A.A., de Azevedo, L.C. (2015) Reduced interhemispheric connectivity in childhood autism detected by electroencephalographic photic driving coherence. *Journal of Autism and Developmental Disorders*, 45(2): 537-547.

Lewis, M., Kim, S.J. (2009) The pathophysiology of restricted repetitive behavior. *Journal of Neurodevelopment Disorders*, 1: 114–132.

Libero, L.E., Maximo, J.O., Deshpande, H.D., Klinger, L.G., Klinger, M.R., Kana, R.K. (2014) The role of mirroring and mentalizing networks in mediating action intentions in autism. *Molecular Autism*, 5: 50. http://www.molecularautism.com/content/5/1/50.

Lotan, M., Shavit, E., Merrick, J. (2015) Enhancing walking ability in individuals with Rett syndrome through the use of applied behavioral analysis (ABA): Review and a case study. *The Open Rehabilitation Journal*, 8: 1-8.

Mendes, D.Q., Carvalho, L.A.V., Wedemann, R.S. (2004) An unifying neuronal model for normal and abnormal thinking. *Learning and Nonlinear Models*, 2(1): 01-13.

Minshew, N.J., Meyer, J., Goldstein, G. (2002) Abstract reasoning in autism: A dissociation between concept information and concept identification. *Neuropsychology*, 16(3): 327-334.

Mittleman, G., Goldowitz, D., Heck, D.H., Blaha, C.D. (2008) Cerebellar modulation of frontal cortex dopamine

efflux in mice: Relevance to autism and schizophrenia. *Synapse*, 62(7): 544-550.

Mostofsky S.H., Ewen, J.B. (2011) Altered connectivity and action model formation in autism is autism. *The Neuroscientist*, 17(4): 437-448.

Muhle, R., Trentacoste, S.V., Rapin, I. (2004) The genetics of autism. *Pediatrics*, 113(5): e472-e486.

Mundy, P. (2003) Annotation: The neural basis of social impairments in autism: The role of the dorsal medial-frontal cortex and anterior cingulated system. *Journal of Child Psychology and Psychiatry*, 44(6): 793-809.

Nakasato, A., Nakatani, Y., Seki, Y., Tsujino, N., Umino, M., Arita, H. (2008) Swim stress exaggerates the hyperactive mesocortical dopamine system in a rodent model of autism. *Brain Research*, 1193: 128-135.

Narita, N., Kato, M., Tazoe, M., Miyazaki, K., Narita, M., Okado, N. (2002) Increased monoamine concentration in the brain and blood of fetal thalidomide- and valproic acid-exposed rat: putative animal models for autism. *Pediatric Research*, 52(4): 576-579.

Nguyen, M., Roth, A., Kyzar, E.J., Poudel, M.K., Wong, K., Stewart, A.M., Kalueff, A.V. (2014) Decoding the contribution of dopaminergic genes and pathways to autism spectrum disorder (ASD). *Neurochemistry International*, 66: 15-26.

Noriega, G. (2007) Self-organizing maps as a model of brain mechanisms potentially linked to autism. *IEEE Transactions on Neural Systems and Rehabilitation Engineering*, 15(2): 217-226.

Noriega, G. (2015) A neural model to study sensory abnormalities and multisensory effects in autism. *IEEE Transactions on Neural Systems and Rehabilitation Engineering*, 23(2): 199–209.

O´Hearn, K., Miya, A., Ordaz, S., Luna, B. (2008) Neurodevelopment and executive function in autism. *Development and Psychopatology*, 20: 1103-1132.

O'Loughlin, C., Thagard, P. (2000) Autism and coherence: A computational model. *Mind & Language*, 15: 375–392.

Paval, D. (2017) A Dopamine Hypothesis of Autism Spectrum Disorder. *Developmental Neuroscience*, 39:355-360.

Passerino, M. Liliana, Santarosa, Lucila C. M., Tarouco, Liane M. R. (2007) Interação social e mediação em ambientes digitais de aprendizagem com sujeitos com autismo. *Revista Brasileira de Informática e Educação*, 15(1): 9-20.

Pelphrey, K.A., Shultz, S., Hudac, C.M., Vander Wyk, B.C. (2011) Research review: Constraining heterogeneity: The social brain and its development in autism spectrum disorder. *The Journal of Child Psychology and Psychiatry*, 52(6): 631–644.

Ritter, H., Kohonen, T. (1989) Self-Organizing Semantic Maps. *Biological Cybernetics*, 61(44): 241-254.

Rogers, S.J., Pennington, B.F. (1991) A theoretical approach to the deficits in infantile autism. *Development and Psychopatology*, 3: 137-162.

Rubenstein, J.L.R., Merzenich, M.M. (2003) Model of autism: increased ratio of excitation/inhibition in key neural systems. *Genes, Brain and Behavior*, 2: 255–267.

Schipul, S.E., Keller, T.A., Just, M.A. (2011) Inter-regional brain communication and its disturbance in autism. *Frontiers in Systems Neuroscience*, 5:1-11.

Servan-Schreiber, D., Printz, H., Cohen, J. (1990) A network model of catecholamine effects: gain, signal-to-noise ratio, and behavior. *Science*, 249: 892-895.

Zandt, F., Prior, M., Kyrios, M. (2007) Repetitive behaviour in children with high functioning autism and obsessive compulsive disorder. *Journal of Autism and Developmental Disorders*, 37: 251-259.

Zikopoulos, B., Barbas, H. (2010) Changes in prefrontal axons may disrupt the network in autism. *The Journal of Neuroscience*, 30(44): 14595-14609.

**FIGURES**

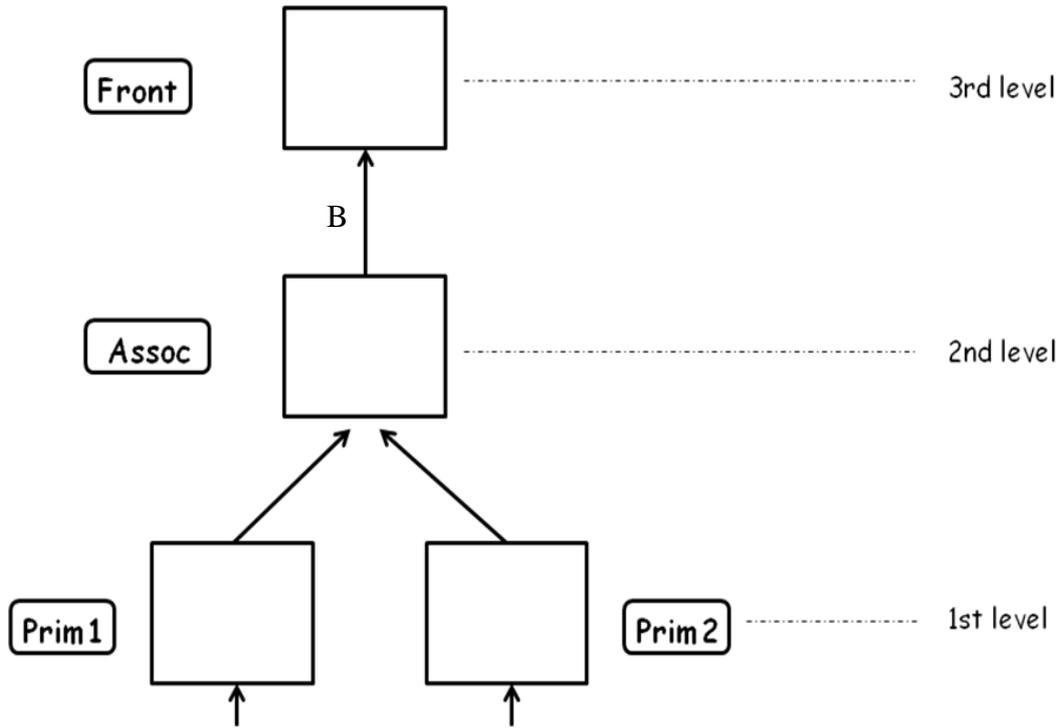

**Figure 1:** The modeled neural network representing a cortical circuit composed by maps hierarchically organized in three levels. Each map corresponds to a distinct cortical area. At the network first level, *Prim1* and *Prim2* represent primary maps. They send projections to a second level map, *Assoc*, which pertains to an associative cortical area. The *Assoc* output reaches the *Front* map, at the network third level. This last map characterizes another associative region - in the frontal cortex.

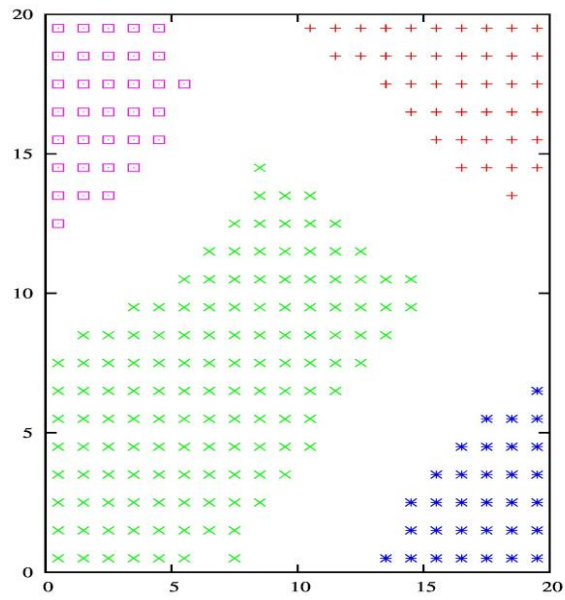

**Figure 2a:** Normal condition *Prim1* map.

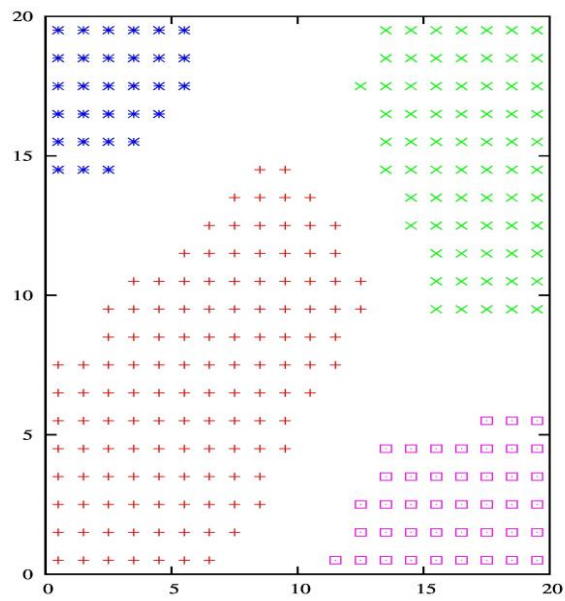

**Figure 2b:** Normal condition *Prim2* map.

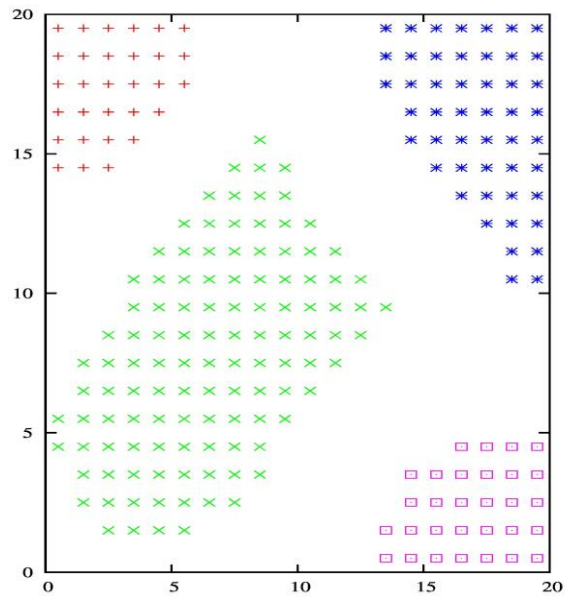

**Figure 2c:** Normal condition *Assoc* map.

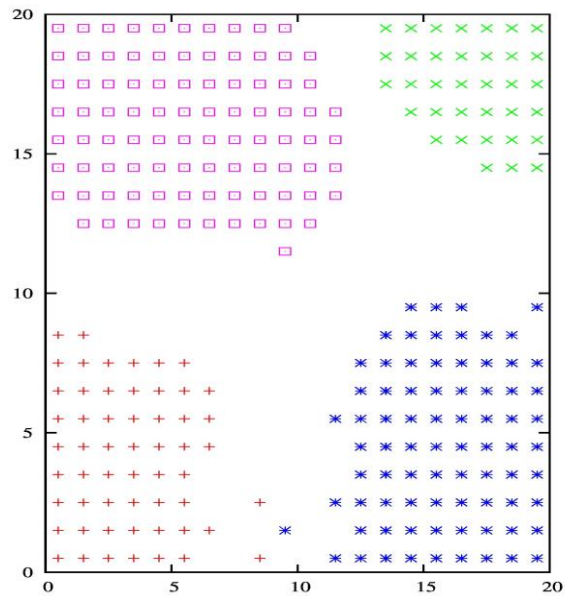

**Figure 2d:** Normal condition *Front* map.

**Figures 2a-d:** Normal condition: *Prim1*, *Prim2*, *Assoc* and *Front* maps, respectively. A referential network, where all maps qualitatively keep the ability of forming categories.

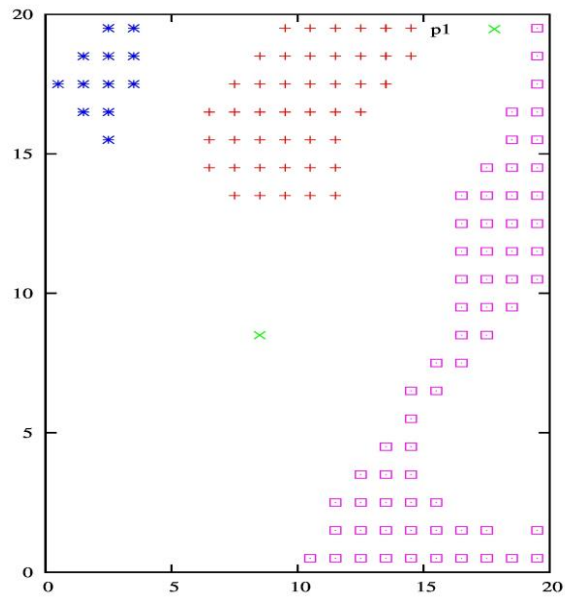

**Figure 3a**

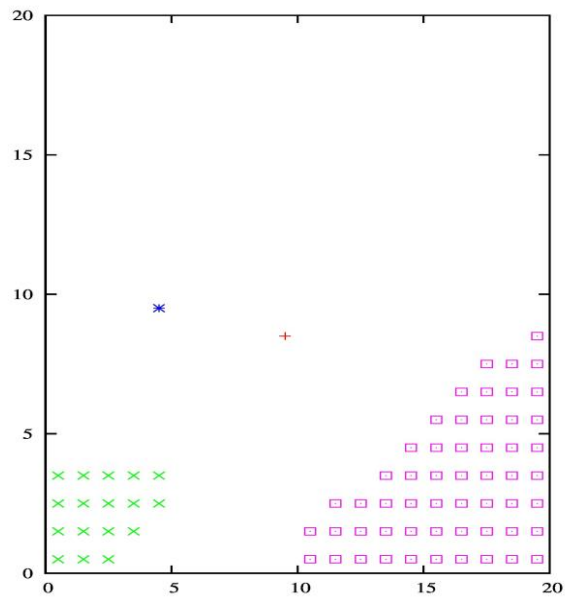

**Figure 3b**

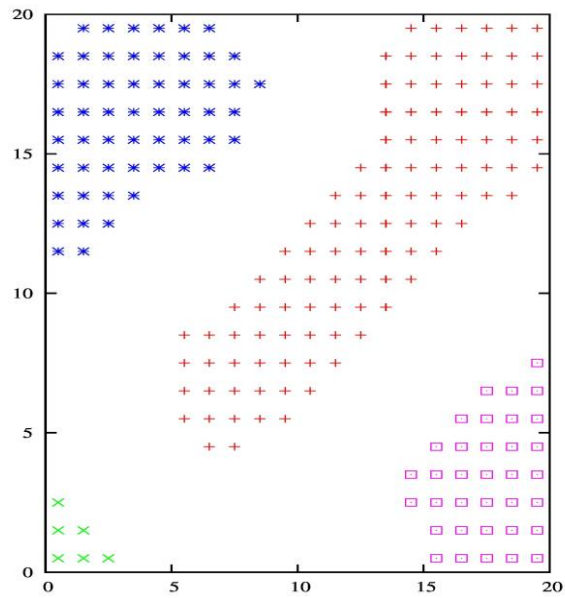

**Figure 3c**

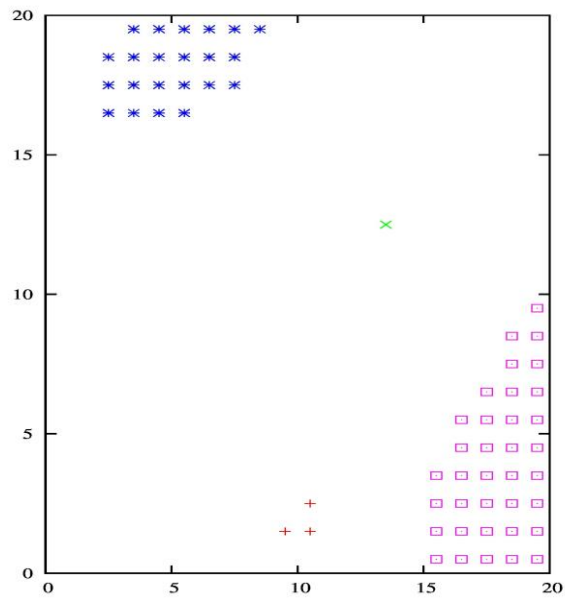

**Figure 3d**

**Figures 3a-d:** Excitatory synapses over-strengthening: when weights suffer an over-updating, with $\Omega = 3$, the map does not achieve a stable state.

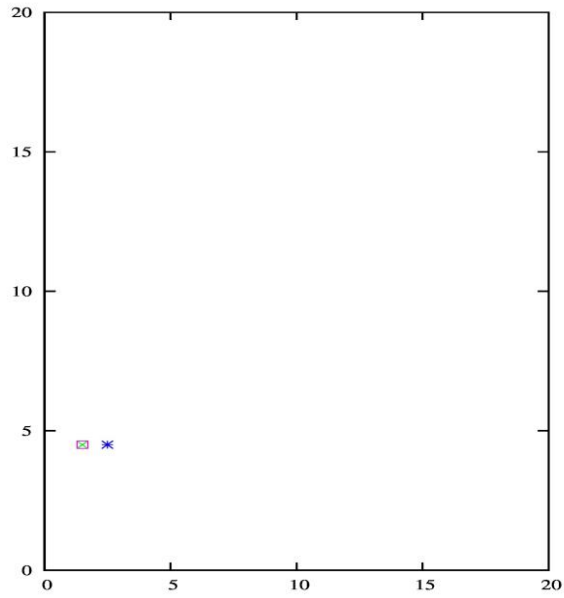

**Figure 4:** Exacerbation of excitatory synapses over-strengthening: no neural structure supports the construction of categories, when Ω = 5, during the weights updating.

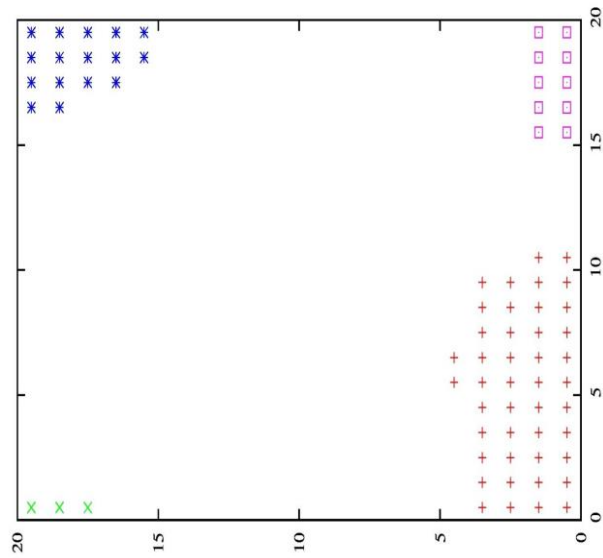

**Figure 5a:** Excitatory area enlargement: σ = 5.0. Very distant clusters and only a few encoding neurons indicating poor, not related concepts.

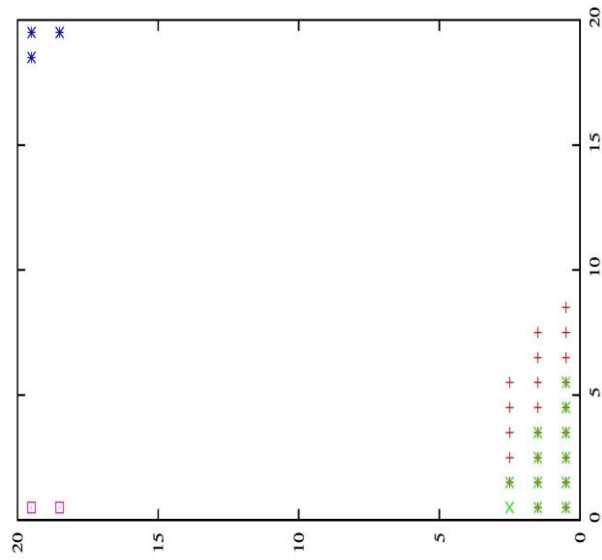

**Figure 5b**

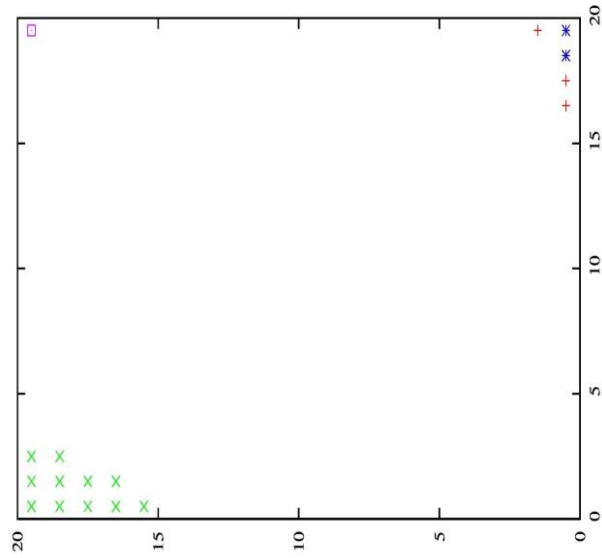

**Figure 5c**

**Figures 5b, c:** Excitatory area enlargement: σ = 7.0 and σ = 8.0, respectively. Enormous distance between clusters, low number of encoding neurons, and superposing of codification: lack of neural structure supporting selectiveness and flexible cognition.

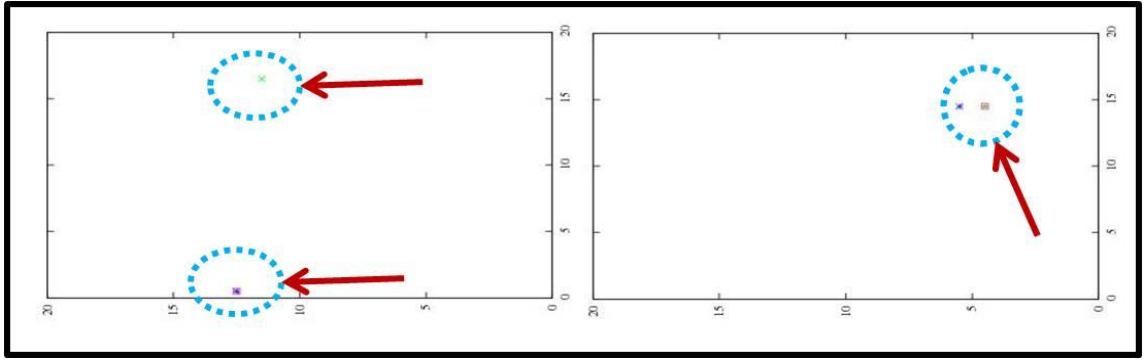

**Figures 6a, b:** Local impairments in maps *Prim1* and *Prim2*: underdevelopment due to over excitatory synaptic strengthening. Minute amount of encoding neurons caused the emergence of similar inputs patterns to *Assoc*.

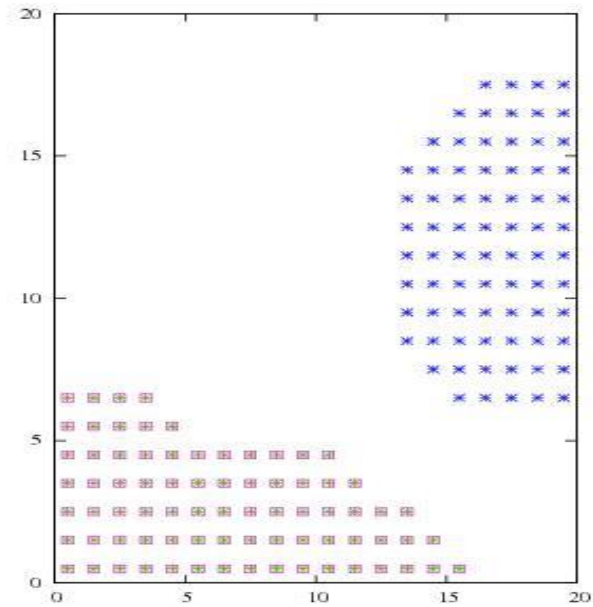

**Figure 6c:** Local impairments influence on higher level map: *Assoc* map development under repetition of fixed inputs. Superposition of codifications and large cortical areas processing restricted signals.

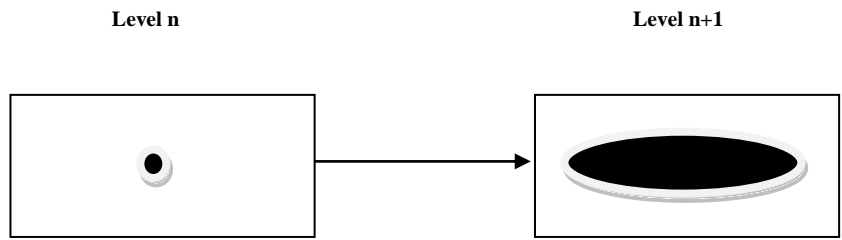

**Figure 6d:** The presence of only a few encoding neurons at the Assoc Map produces steady and repetitive signals that lead to the development of a unique, exacerbate neurons cluster at the Front Map to represent them.

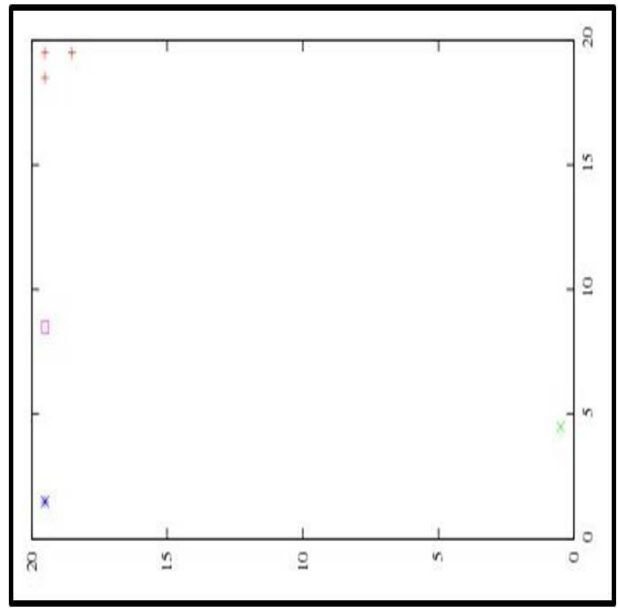

**Figure 7a:** Long connection 'A' disruption, ρ = 0.02: few isolated neurons in *Assoc* map respond to afferent signals, thus generating fixed, repetitive inputs to the third level map.

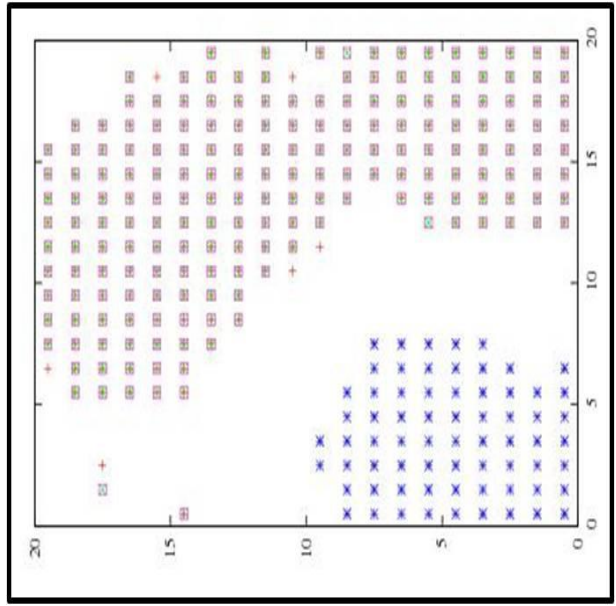

**Figure 7b:** Long connection disruption influence on *Front* map. Distinct classes of stimuli are encoded by the same neuron cluster, which forms a markedly large processing region throughout the *Front* map: lack of selectivity besides predominance of fixed signals processing.

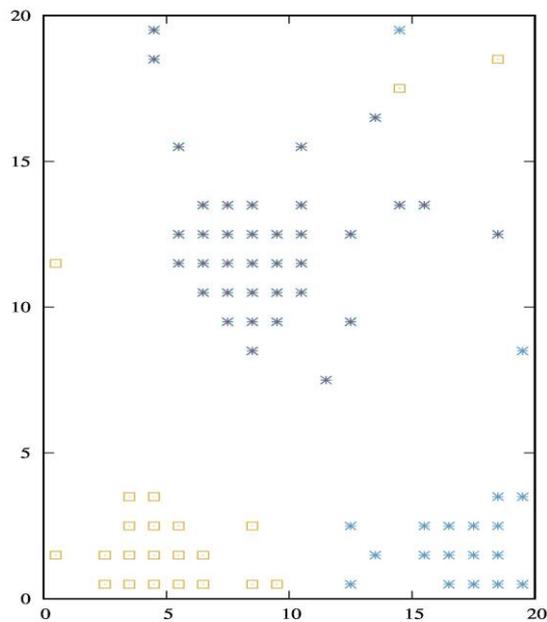

**Figure 8:** Long connection 'B' disruption, ρ = 0.02: a non consolidated *Front* map, with loss of representativeness and selectivity.

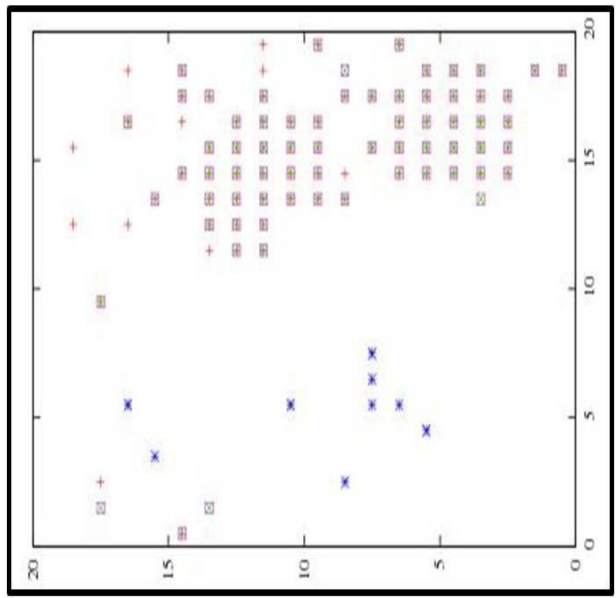

**Figure 9:** Long connections 'A' and 'B' disruptions, ρ = 0.02: a even more spoilt, non consolidated *Front* map.

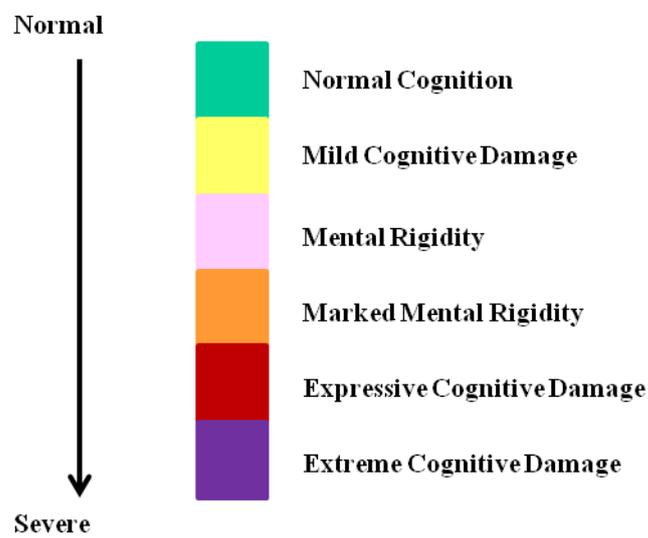

**Figure 10a:** Putative cognitive symptoms.

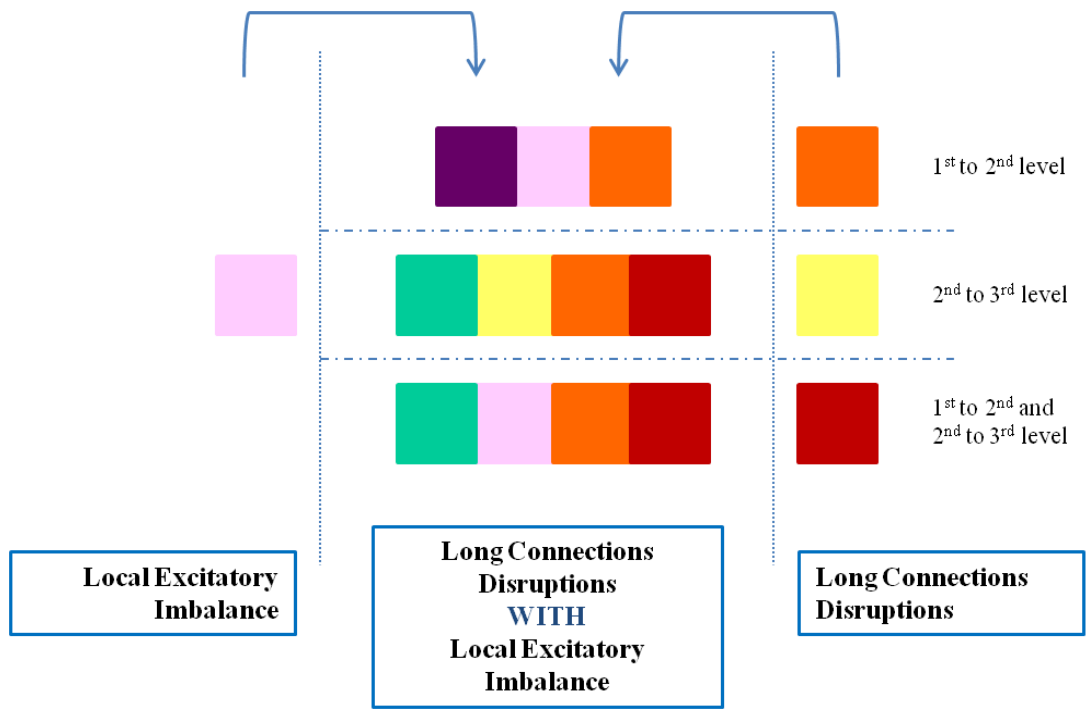

**Figure 10b:** Putative cognitive symptoms resulting from excitatory and/or connectivity alterations.

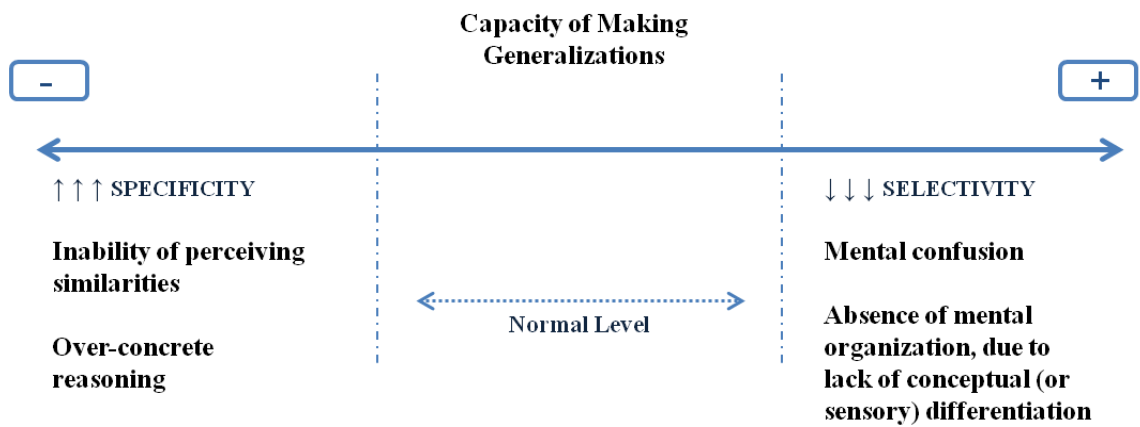

**Figure 11:** Making generalizations: a *continuum* between over-specificity and under-selectivity.